\documentclass[runningheads]{llncs}
\usepackage{times}
\usepackage{graphicx}
\usepackage{amsmath,amssymb} 
\usepackage{color}
\usepackage[width=122mm,left=12mm,paperwidth=146mm,height=193mm,top=12mm,paperheight=217mm]{geometry}
\usepackage{multirow}
\usepackage{placeins}

\begin{document}
\mainmatter
\title{Deforming Autoencoders: Unsupervised Disentangling of Shape and Appearance} 

\titlerunning{Deforming Autoencoders: Unsupervised Disentangling of Shape and Appearance}

\authorrunning{Shu, Sahasrabudhe, Guler, Samaras, Paragios, Kokkinos.}
\author{Zhixin Shu\textsuperscript{1}
\hspace{0.7cm}
Mihir Sahasrabudhe\textsuperscript{2} 
\hspace{0.7cm}
Alp Guler\textsuperscript{2,3}, \\ 
Dimitris Samaras\textsuperscript{1}
\hspace{0.7cm}
Nikos Paragios\textsuperscript{4} 
\hspace{0.7cm}
Iasonas Kokkinos\textsuperscript{5,6}}


\institute{\textsuperscript{1}Stony Brook University
    \hspace{0.7cm}
	\textsuperscript{2}CVN, CentraleSupélec \hspace{0.7cm}
	\textsuperscript{3}INRIA\\
	\textsuperscript{4}Therapanacea
	\hspace{0.3cm}
	\textsuperscript{5}University College London
    \hspace{0.3cm}
	\textsuperscript{6}Facebook AI Research
}

\newcommand\nl\newline

\maketitle

\begin{abstract}
In this work we introduce  Deforming Autoencoders, a generative model for images that disentangles shape from appearance in an  unsupervised manner.
As in the deformable template paradigm, shape is represented as a deformation between a canonical coordinate system (`template') and an observed image, while appearance is modeled in `canonical', template, coordinates, thus discarding variability due to deformations.
We introduce novel techniques that allow this approach to be deployed in the setting of autoencoders and show that this method can be used for unsupervised group-wise image alignment. We show experiments with expression morphing in humans, hands, and digits, face manipulation, such as shape and appearance interpolation,  as well as unsupervised landmark localization. A more powerful form of unsupervised disentangling becomes possible in template coordinates, allowing us to successfully decompose face images into shading and albedo, and further manipulate face images.
\end{abstract}

\newcommand{\comment}[1]{}
\newcommand{\mycomment}[1]{}

\newcommand{\reffig}[1]{Fig.~\ref{#1}}

\begin{figure}[ht!]
    \vspace{-1.1 cm}
    \centering
    \includegraphics[width=.9\linewidth]{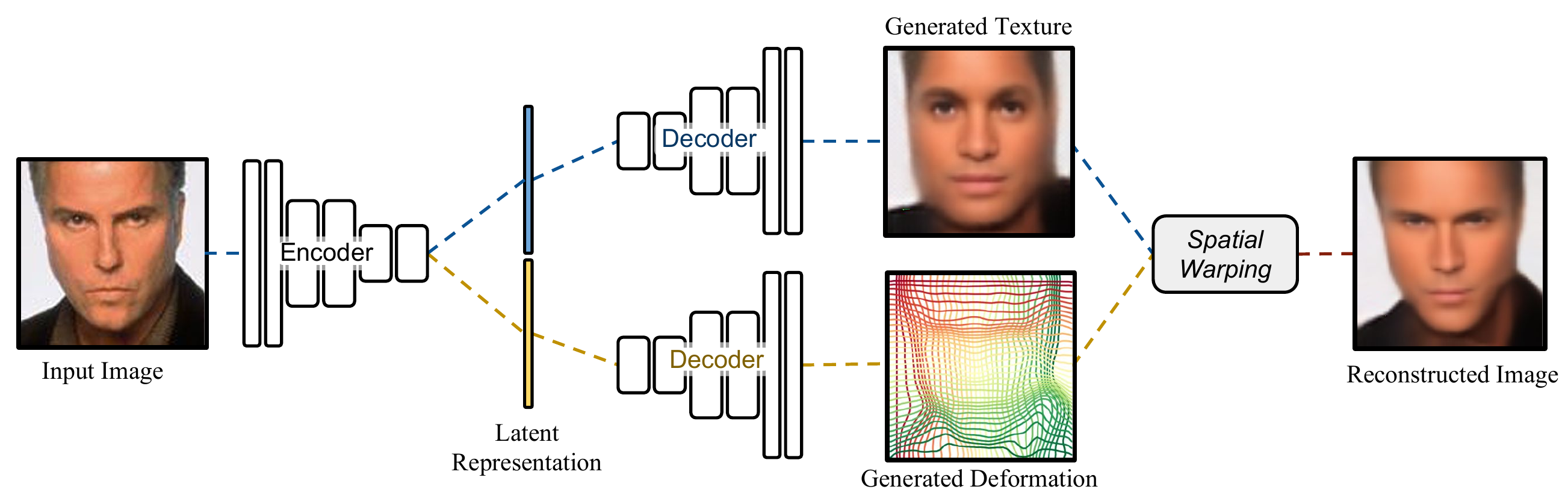}
    \caption{ Deforming Autoencoders follow the deformable template paradigm and model image generation through a cascade of appearance (or, `texture') synthesis in a canonical coordinate system and a spatial deformation that warps the texture to the observed image coordinates. By keeping the latent vector for  texture  short the network is forced to model shape variability through the deformation branch, so as to minimize a reconstruction loss.
    This allows us to train  a deep generative image model that disentangles shape and appearance in an entirely unsupervised manner.}
    \label{fig:teaser}
     \vspace{-0.8cm}
\end{figure}

\section{Introduction}
Disentangling factors of variation is important for the broader goal of 
controlling and understanding deep networks, but also for applications such as image manipulation through interpretable operations.
Progress  in the direction of disentangling the latent space of deep generative models has  facilitated the separation of  latent image representations into dimensions that account for independent factors of variation, such as identity, illumination, normals, and spatial support \cite{infogan,ShuYHSSS17,brostow17,sengupta2017sfsnet}, low-dimensional  transformations, such as rotations, translation, or scaling, 
\cite{hinton10,WorrallGTB16,park2017transformation}
or finer-levels of variation, including age, gender, wearing glasses, or other attributes e.g. \cite{ShuYHSSS17,fader} for particular classes, such as faces. 

Shape variation is more challenging as it amounts to a transformation of a function's domain, rather than its values. Even simple, supervised additive models of shape  result in complex nonlinear optimization problems \cite{cootes1998active,MaBa04}. Despite this challenge
several works in the previous decade aimed at learning shape/appearance factorizations in an unsupervised manner, exploring groupwise image alignment, \cite{congealing,iccv07,FreyJ03,epitome}. In the context of deep learning several works have aimed at incorporating deformations and alignment in a supervised setting, including Spatial Transformers \cite{JaderbergSZK15}, Deep Epitomic Networks \cite{PapandreouKS15}, Deformable CNNs \cite{DaiQXLZHW17},  Mass Displacement Networks \cite{neverova}, Mnemonic Descent \cite{trigeorgis2016mnemonic}, or Densereg \cite{guler2016densereg}. These works have shown that  one can  improve the accuracy of both classification and localization tasks by injecting deformations and  alignment within traditional CNN architectures. 

Turning to unsupervised deep learning, even though most works focus on rigid, or low-dimensional parametric deformations, e.g. \cite{hinton10,WorrallGTB16},
several works have attempted to incorporate richer non-rigid deformations within learning. A thread of works has been aimed at dynamically rerouting the processing of information within the network's graph based on the input, starting from neural computation arguments
\cite{Hinton81,OlshausenAE95,vdm81}
and eventually translating into concrete algorithms, such as the `capsule' works of \cite{Hinton11,Hinton17} that bind neurons on-the-fly. Still, these works lack a transparent, parametric handling of non-rigid deformations. 
Working on a more geometric direction, several works have recently aimed at recovering dense correspondences between pairs \cite{BristowVL15} or sets of RGB images, as e.g. in the recent works of  \cite{ZhouKAHE16,denseiccv17}.
These works however do not have the notion of
a reference coordinate system (`template') to which images can get mapped - this makes the image generation and manipulation harder. More recently, \cite{ThewlisBV17a} use the equivariance principle in order to align sets of images to a common coordinate system, but do not develop this into a full-blown generative model of images.

Our work pushes the envelope of this line of research by following the deformable template paradigm \cite{Grenander1991,yuille1991deformable,cootes1998active,BlVe03,MaBa04}. In particular, we consider that object instances are obtained by deforming a prototypical object, or `template', through dense, diffeomorphic deformation fields.  
This makes it possible to factor object variability within a category into variations that are associated to  spatial transformations, generally linked to the object's 2D/3D shape, and variations that are associated to appearance (or, `texture' in graphics), e.g. due to facial hair, skin color, or illumination. 
In particular we consider that both sources of variation can be modelled in terms of a low-dimensional latent code that is learnable in an unsupervised manner from images. We achieve disentangling by breaking this latent code into separate parts that are fed into separate decoder networks that deliver appearance and deformation estimates. Even though one could hope that a generic convolutional architecture will learn to represent such effects, we argue that explicitly injecting this inductive bias in a network can help with the  training, while also yielding control over the generative process.

Our main contributions in this work can be summarized as follows:

\newcommand{\dae}{Deforming~~Autoencoder}
First, we introduce the {\em{\dae}} architecture, bringing together the deformable modeling paradigm with unsupervised deep learning. We treat the template-to-image correspondence task as that of predicting a smooth and invertible transformation. As shown in \reffig{fig:teaser}, our network predicts this transformation field  alongside with the template-aligned appearance and subsequently deforms the synthesized appearance to generate an image similar  to its input. This allows for a disentanglement of the shape and appearance parts of image generation by explicitly modelling the effects of image deformation during the decoding stage.

Second, we explore different ways in which deformations can be represented and predicted by the decoder. 
Instead of building a generic deformation model, 
we compose a global, affine deformation field, with a non-rigid field that is synthesized as a convolutional decoder network. 
We develop a method that allows us to 
constrain the synthesized field to be a diffeomorphism, namely an invertible and smooth transformation, and show that it  simplifies training and improves accuracy. 
We also show that class-related information can be exploited, when available, to learn better deformation models: this yields sharper images and can be used to learn models that jointly account for multiple classes - e.g. all MNIST digits.  

Third, we show that disentangling appearance from deformation comes with several advantages when it comes to modeling and manipulating  images. By using disentangling we obtain  
clearly better synthesis results when manipulating images for tasks such as expression, pose or identity interpolation  when compared to standard  autoencoder architectures.
Along the same lines, we show that accounting for deformations facilitates a further  disentangling of the appearance components into an intrinsic, shading-albedo decomposition which completely fails when naively performed in the original image coordinates. This allows us to perform re-shading through simple operations on the latent shading coordinate space.

We complement these qualitative results with a quantitative analysis of the learned model in terms of  landmark localization accuracy. We show that our method is not too far below supervised methods and outperforms with a margin the latest state-of-the-art works on self-supervised correspondence estimation \cite{ThewlisBV17a}, even though we never explicitly trained our network for correspondence estimation, but rather only aimed at reconstructing pixel intensities.  

\section{Deforming Autoencoders}

\newcommand{\obs}{I} 
\newcommand{\shp}{W}
\newcommand{\temp}{T} 
\newcommand{\xj}{\mathbf{p}}
\newcommand{\xy}{(x,y)}
\newcommand{\shb}{W} 
\newcommand{\beq}{\begin{equation}}
\newcommand{\eeq}{\end{equation}}
\newcommand{\ba}{\begin{equation*}}
\newcommand{\ea}{\end{equation*}}

Our architecture embodies the deformable template  paradigm in an autoencoder architecture. The  premise of our work is that 
image generation can be interpreted as the combination of two processes: a synthesis of appearance on a deformation-free coordinate system (`template'), followed by a subsequent deformation that introduces shape variability. 
Denoting by $\temp(\xj)$ the value of the synthesized appearance (or, texture) at coordinate $\xj = \xy$ and by $\shb(\xj)$ the estimated deformation field, we consider that the observed image, $\obs(\xj)$, can be reconstructed as follows:
\beq
\obs(\xj) \simeq \temp(\shp(\xj)),
\eeq 
namely  the image appearance at position $\xj$ is obtained by looking up the synthesized appearance at position $\shp(\xj)$. This is implemented in terms of a spatial transformer layer \cite{JaderbergSZK15} that allows us to pass gradients through the warping process.

The appearance and deformation functions are synthesized
by independent decoder networks. The inputs to the
decoders are delivered by a joint encoder network that takes
as input the observed image and delivers a low-dimensional
latent representation, $Z$, of shape and appearance. This is
split into two parts, $Z = [Z_T, Z_S]$ which feed into the appearance and shape networks respectively, providing us with a clear separation of shape and appearance.

\subsection{Deformation field modeling}
Rather than leave deformation modeling entirely to back-propagation, we use some domain knowledge to simplify and accelerate learning. The first  observation is that global aspects can be expressed using low-dimensional linear models. We account for global deformations by an affine Spatial Transformer layer, that uses a six-dimensional input to synthesize a deformation field as an expansion on a fixed basis \cite{JaderbergSZK15}.
This means that the shape representation, $Z_S$ described above is decomposed into two parts, $Z_W,Z_A$, where $Z_A$ accounts for the affine, and $Z_W$ for the non-rigid, learned part of the deformation field. 
These deformation fields are generated by separate decoders, and are {\em composed}, so that the affine transformation warps the detailed non-rigid warps to the image positions where they should apply. This is also a common decomposition in deformable models for faces \cite{cootes1998active,MaBa04}. 

\begin{figure}
    \centering
    \includegraphics[width=0.99\linewidth]{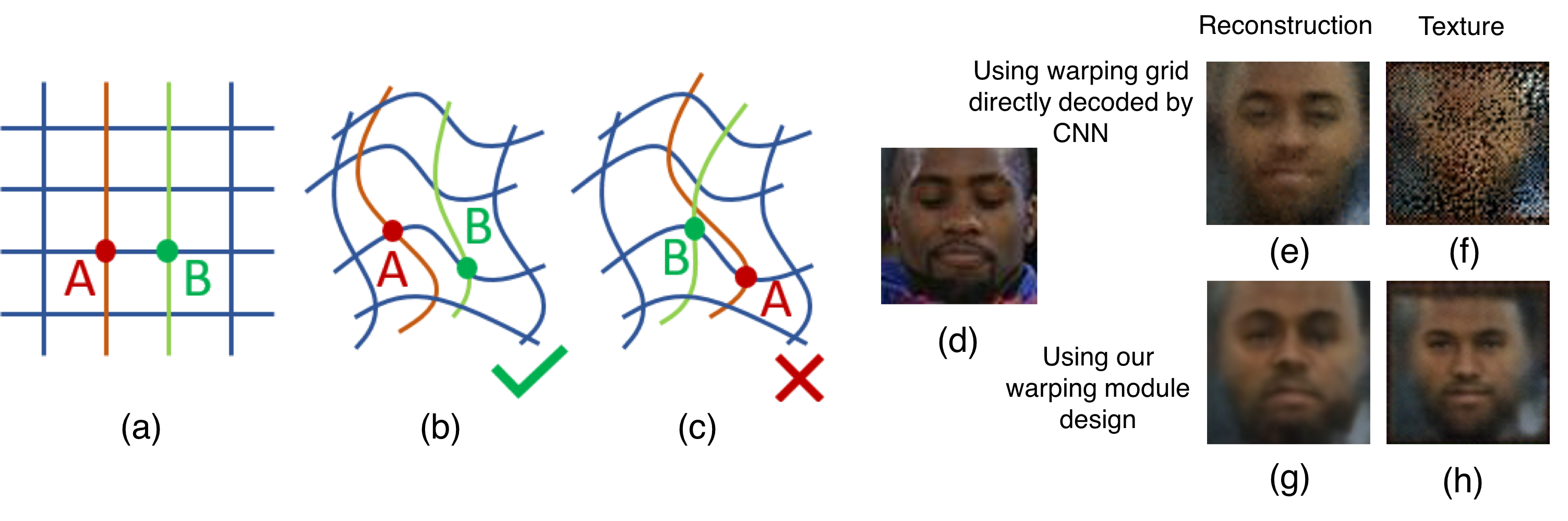}
    \caption{Our warping module design only permits locally consistent warping, as shown in (b), while the flipping of  relative pixel positions, as shown in (c), is not allowed by design. 
    To achieve this, we  let the deformation decoder predict the horizontal and vertical increments of the deformation ($\nabla_x W$ and $\nabla_y W$, respectively) and use a ReLU transfer function to remove local flips, caused by going back in the vertical or horizontal direction. A spatial integral module is subsequently applied to generate the grid. This simple mechanism
    serves as an effective constraint for the deformation generation process, while allowing us to model free-form/non-rigid local deformation. }
    \label{fig:gradient}
     \vspace{-0.3cm}
\end{figure}

Turning to local deformation effects, we quickly realized that not every deformation field is plausible. Without appropriate regularization we would often obtain deformation fields that could expand small areas to occupy whole regions, and/or would be non-diffeomorphic, meaning that the deformation could spread a connected texture pattern to a disconnected image area (Figure~\ref{fig:gradient}-(f)). 

\label{integral}
To prevent this problem, instead of making the shape decoder CNN directly predict the local warping field $\shp(\xj) = (\shp_x(x,y),\shp_y(x,y))$, we consider a `differential decoder' that generates the spatial gradient of the warping field: $\nabla_x \shp_x$ and $\nabla_y \shp_y$, where $\nabla_{c}$ denotes the $c-th$ component of the spatial gradient vector.   These two quantities measure the displacement of consecutive pixels -  for instance  $\nabla_x \shp_x=1$ amounts to translation in the horizontal axis, $\nabla_x \shp_x=2$ amounts to horizontal shifting by a size of 2, while $\nabla_x \shp_x=-1$ amounts to left-right flipping;  a similar behavior is associated with $\nabla_y \shp_y$ in the vertical axis.   We note that global rotations are handled by the affine warping field, and the $\nabla_x \shp_y, \nabla_y \shp_x$ are associated with small local rotations of minor importance - we therefore focus on $\nabla_x \shp_x, \nabla_y \shp_y$.

Having access to these two values gives us a handle on the deformation field, since we can prevent folding/excessive stretching by controlling $\nabla_x \shp_x, \nabla_y \shp_y$. 

In particular, we pass the outputs of our differential decoder through a Rectified Linear Unit (ReLU) module, which enforces positive horizontal offsets on horizontally adjacent pixels,
and positive vertical offsets on vertically adjacent pixels. 
We subsequently apply a spatial integration layer, implemented in terms of a fixed network layer, on top of the output of the ReLU layer to reconstruct the warping field from its spatial gradient. By doing so, the new deformation module enforces the generation of smooth and regular warping fields that avoid self-crossings.  
In practice we found that also clipping  the decoded offsets by a maximal value significantly eases the training, which amounts to replacing the ReLU layer, $\mathrm{ReLU}(x) = \max(x,0)$ with a  $\mathrm{HardTanh}_{0,\delta}(x) = \min(\max(x,0,\delta)$ layer. In our experiments, we set $\delta = 5/w$ where $w$ denotes the number of pixels along one dimension of the image.

\subsection{Class-aware Deforming Autoencoder}
\label{section:class-aware}
\label{class}
\begin{figure}[ht!]
 \vspace{-0.25cm}
    \centering
    \includegraphics[width=0.7\linewidth]{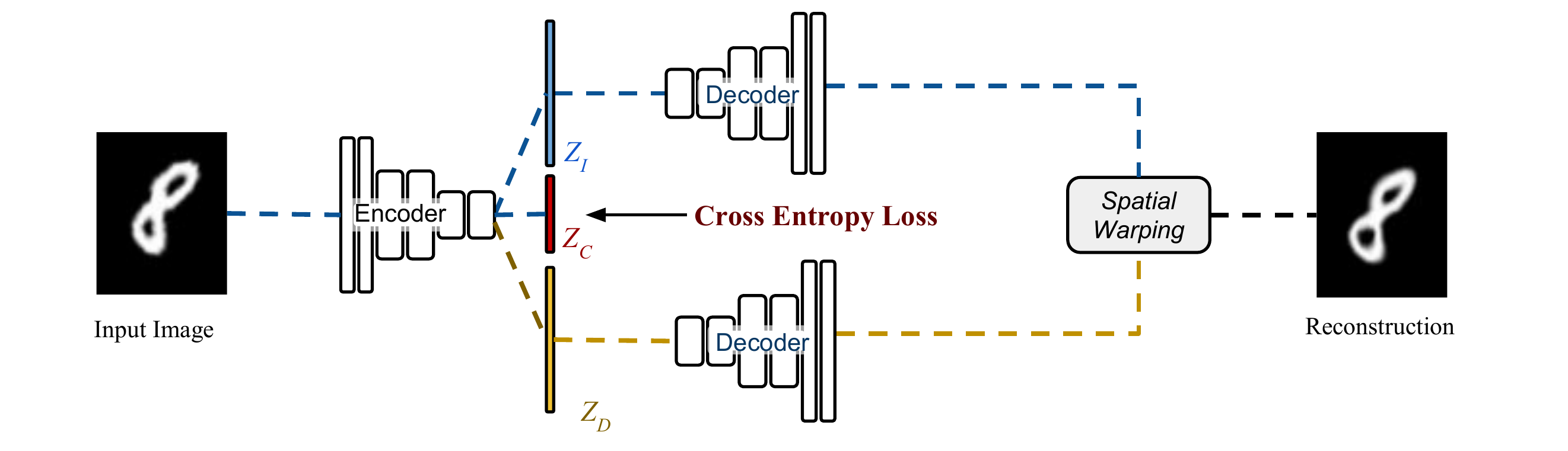}
    \caption{A \emph{class-aware} model can account for multi-modal   deformation distributions by utilizing class
        information. Introducing a classification loss into latent space
        helps the model learn a better representation of the input as demonstrated on MNIST.}
    \label{fig:EncDecInject}
\end{figure}

 We can require our network's latent representation to be predictive of not only shape and appearance, but also of instance class, if that is available during training. We note that this information, being discrete may be  easier to acquire than the actual deformation field, which would require manual landmark annotation. For instance, for faces such discrete information could represent the expression or a person's identity. 
 
In particular we consider that the latent representation can be decomposed as follows:
$Z = [Z_T,Z_C,Z_S]$, where $Z_T,Z_S$ are as previously the appearance- and shape- related parts of the representation, respectively, while $Z_C$ is fed as input to a sub-network  trained to predict the class associated with the input image. Apart from assisting the classification task, the latent vector $Z_C$ is fed into both the appearance and shape decoders. 
Intuitively this allows our decoder network to learn a mixture model that is conditioned on class information,
rather than treating the joint, multi-modal distribution through a monolithic model. Even though the class label is only used during training, and not for reconstruction, our experimental results show that a network trained with class supervision can deliver more accurate synthesis results.

\subsection{Intrinsic Deforming Autoencoder: Deformation, Albedo and Shading  Decomposition}
\label{sec:idae}
\begin{figure}[h]
    \centering
    \includegraphics[width=0.99\linewidth]{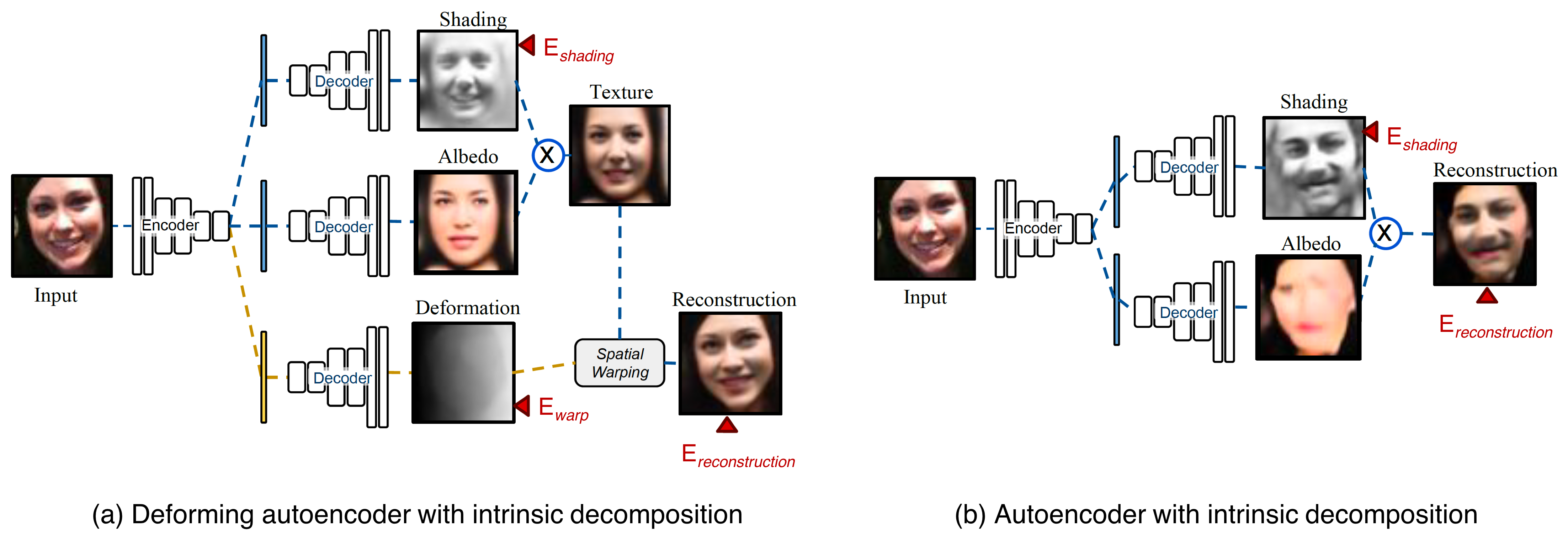}
    \caption{Autoencoders with intrinsic decomposition. (a) Deforming Autoencoder with intrinsic decomposition (Intrinsic-DAE): we model the texture by the Hadamard product of shading and albedo components, each of which is decoded by an individual decoder. The texture is subsequently warped by the predicted deformation field. (b) A plain autoencoder with intrinsic decomposition. Both networks are trained with reconstruction loss ($E_{\text{Reconstruction}}$) on the final output and regularization losses on shading ($E_{\text{Shade}}$) and deformation ($E_{\text{Warp}}$), if it exists. }
    \label{fig:daeIntrinsicStructure}
\end{figure}

Having outlined  Deforming Autoencoders, we now use a Deforming Autoencoder to  model complex physical image signals, such as illumination effects, without a supervision signal. For this we design the Intrinsic Deforming-Autoencoder, named Intrinsic-DAE to model shading and albedo for in-the-wild face images. As shown in Fig.~\ref{fig:daeIntrinsicStructure}-(a), we introduce two separate decoders for shading $S$ and albedo $A$, each of with has the same structure as the original texture decoder. The texture is computed by $T=S \circ A$ where $\circ$ denotes the Hadamard product. 

In order to model the physical properties of shading and albedo, we follow the intrinsic decomposition regularization loss used in~\cite{ShuYHSSS17}: we apply the L2 smoothness loss on $\nabla S$, meaning that shading is expected to be smooth, while leaving albedo unconstrained. As shown in Fig.~\ref{fig:daeIntrinsicStructure} and more extensively in the experimental results section, when used in tandem with an Deforming Autoencoder this allows us to successfully decompose of face image into shape, albedo, and shading components, while a standard Autoencoder completely fails at decomposing unaligned images into shading and albedo.   

\subsection{Training}

Our objective function is formed as the sum of three losses, combining the reconstruction error with the regularization terms required for the modules described above. Concretely, the loss of the deforming autoencoder can be written as
\begin{equation}
E_{\text{DAE}} = E_{\text{Reconstruction}} + E_{\text{Warp}},
\end{equation}
where the reconstruction loss is defined as the standard $\ell_2$ loss
\begin{equation}
E_{\text{Reconstruction}} = \|I_{\text{Output}} - I_{\text{Input}}\|^2,
\end{equation}
and the warping loss is decomposed as follows:
\begin{equation}
E_{\text{Warp}} = E_{\text{Smooth}} + E_{\text{BiasReduce}}.
\end{equation}
In particular the smoothness cost, $E_{\text{smooth}}$, penalizes quickly-changing deformations encoded by the local warping field. It is measured in terms of the total variation 
norm of the horizontal and vertical differential warping fields, and is given by
\begin{equation}
E_\text{Smooth} =  \lambda_1 \left(\| \nabla W_{x}(x,y)\|_1 + \| \nabla W_{y}(x,y)\|_1\right),
\end{equation}
where $\lambda_1=1e-6$.
Finally, $E_{\text{BiasReduce}}$ aims at removing any systematic bias introduced by the fitting process, e.g. the average template becoming small, or a distorted version of the data. It consists of regularization
on (1) the affine parameters defined as the L2-distance between $S_A$ and $S_0$,  with
$S_0$ being the identity affine transform and (2) on free-form deformations defined as the L2-distance between the average deformation grid within a minibatch, $\bar{W}$ and the identity grid $W_0$:
\begin{equation}
E_{\text{BiasReduce}} = \lambda_2 \|S_A - S_0\|^2 + \lambda_2' \|\bar{W} - W_0\|^2,
\end{equation}
where $\lambda_2=\lambda_2'=0.01$. 

In the class-aware variant described in Sec.~\ref{class} we augment the loss above with the cross-entropy loss evaluated on the classification network's outputs, while for Intrinsic-DAE, we add the following objective function in training:\\
\noindent $ E_{\text{Shade}} = \lambda_3 \|\nabla S\|^2$ where $\lambda_3 = $1e-6.

We experiment with two types of architectures; the majority of our results are obtained with a standard auto-encoder architecture, 
where both encoder and decoders are CNNs with standard convolution-BatchNorm-ReLU blocks. The number of filters and the texture bottleneck capacity can vary per experiment, image resolution, and dataset, as  detailed in the Appendix~\ref{append:architecture}. 

Follow the recent work on densely connected convolutional networks \cite{huang2017densely}, we have also experimented with incorporating dense connections into our encoder and decoders architectures respectively (no skip connections over the bottleneck layer for latent representations). In particular, we follow the architecture of DenseNet-121, but without the $1 \times 1$ convolutional layers inside each dense block. These have been shown to better exploit larger datasets, as indicated in the quantitative analysis of unsupervised face alignment. We call this version of the deforming autoencoder Dense-DAE.

\section{Experiments}

\begin{figure}
    \centering
    \includegraphics[width=0.95\linewidth]{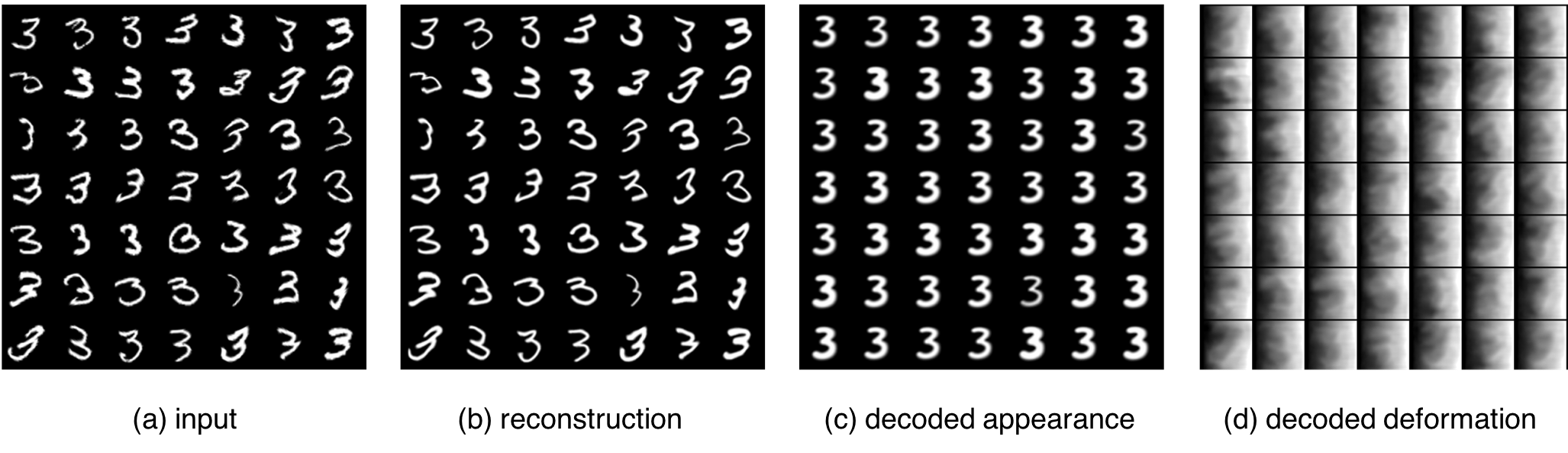}
    \caption{Unsupervised deformation-appearance disentangling on a single MNIST digit. Our network learns to reconstruct the input image while automatically deriving a canonical appearance for the input image class. In this experiment, the dimension of the latent representation for appearance $Z_T$ is 1.}
    \label{fig:mnist_1}
\end{figure}

To demonstrate the properties of our deformation disentangling network, we conduct experiments on the following three datasets:
\begin{itemize}
\item Deformed MNIST.  ~~A synthetic dataset designed specifically to explore the deformation modelling power of our network. Deformed MNIST consists of 
      handwritten MNIST images randomly distorted using a mixture of sinusoidal waveforms. 
\item MUG facial expression dataset \cite{mugfacialexpr}. ~~This dataset consists of videos of individuals performing facial expressions, with simple blue background and minor translation. The dataset also offers frames from the videos, classified according to the facial expression, as well as the subject.
\item Faces-in-the-wild dataset: MAFL \cite{xiaooutangmafldataset} and CelebA \cite{liu2015faceattributes}. These datasets consist of uncontrolled ``in-the-wild'' faces with variability in  pose, illumination, expression, age, etc.
\end{itemize}

Using these datasets we  experimentally explored the ability of the unsupervised appearance-shape (or texture-deformation) disentangling network on 1) unsupervised image alignment/appearance inference; 2) learning semantically meaningful manifolds for shape and appearance; 3) decomposition into illumination intrinsics (shading, albedo); 4) unsupervised landmark detection, as detailed below. We intend to make all of the code of our system publicly available in order to facilitate the reproduction of our results. 

\subsection{Unsupervised Appearance Inference}

\begin{figure}
    \centering
    \includegraphics[width=0.99\linewidth]{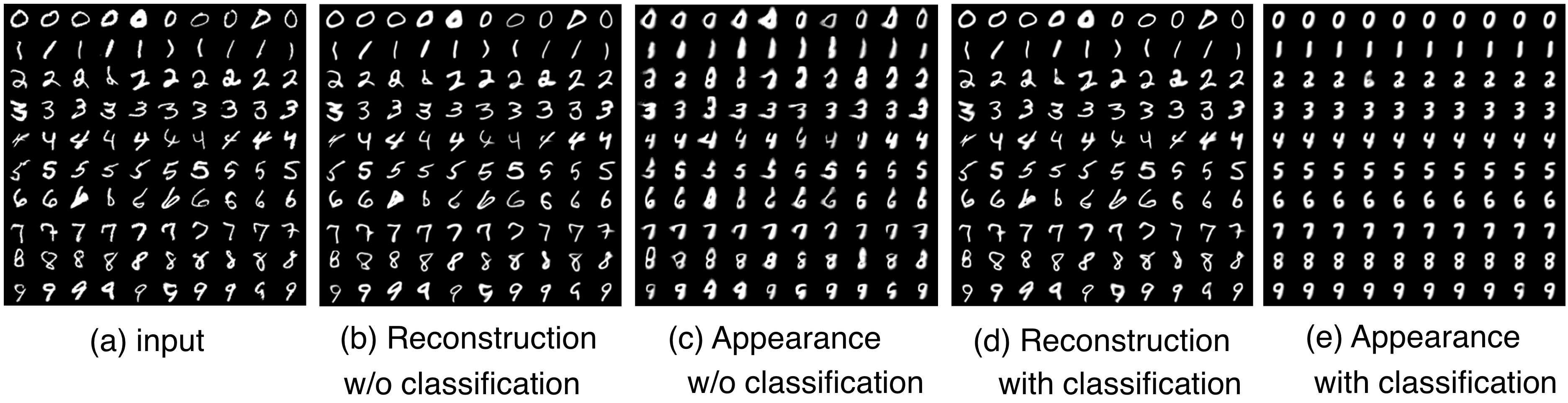}
    \caption{Class-aware Deforming Autoencoders effectively model the appearance and deformation for multi-class data.}
    \label{fig:mnist_all}
\end{figure}

\begin{figure}
    \centering
    \includegraphics[height=0.33\linewidth]{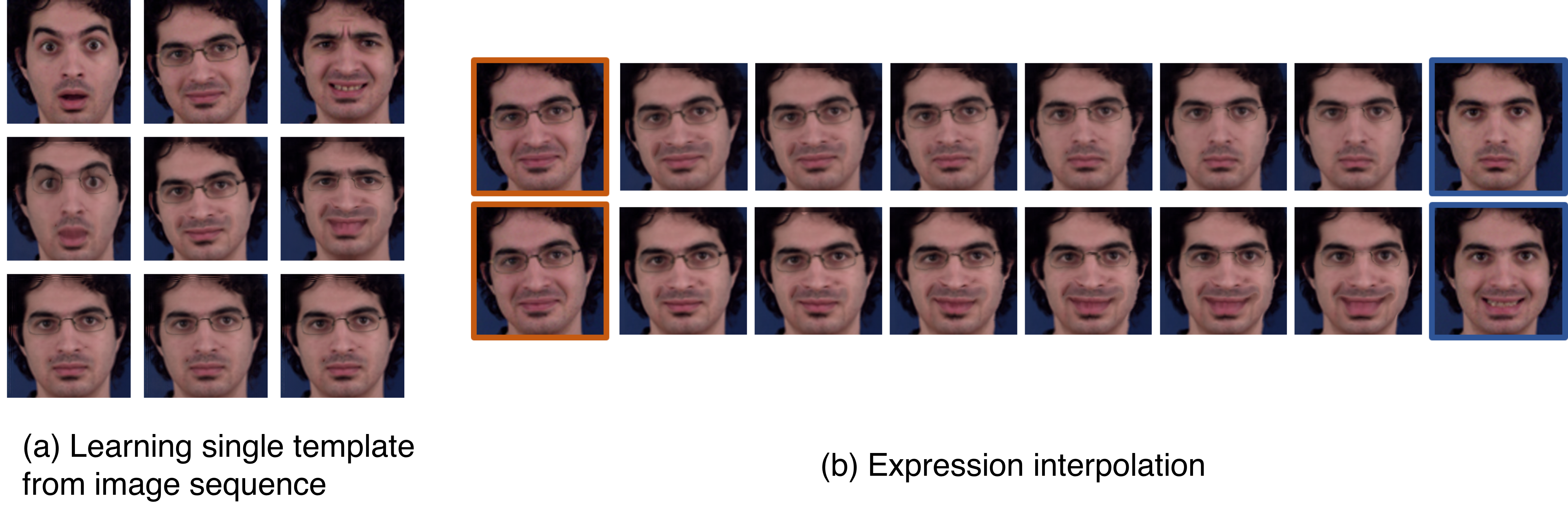}
    \caption{ Experiment on MUG dataset of face expressions: (a) With 0-length $Z_T$, Deforming Autoencoders learn a single texture (row 3) from a subject in the MUG facial expression dataset. By doing so, the subject's facial expression is encoded only in the deformation domain.
    (b): Our network is able to disentangle the facial expression deformation and encode this information in a meaningful latent representation. By interpolating the latent deformation representation from the source (in orange) to the target (in blue), it generates sharp images and a smooth deformation interpolation between expressions as shown in each row.
    }
    \label{fig:MUG_1}
\end{figure}

We first use our network to model canonical appearance and deformation for single category objects. For this purpose, we demonstrate the results in the MNIST and MUG facial expression datasets (Fig.~\ref{fig:mnist_1},~\ref{fig:mnist_all},~\ref{fig:MUG_1}).

We observe that by heavily limiting the size of $Z_T$ (1 in Fig.~\ref{fig:mnist_1} and 0 in Fig.~\ref{fig:MUG_1}), we can successfully infer a canonical appearance for such a class. In Fig.~\ref{fig:mnist_1}, all different types of handwritten digits '3' are aligned to a simple canonical shape. In Fig.~\ref{fig:MUG_1}, by limiting the dimension of $Z_T$ to $0$, the network learns to encode a single texture image for all expressions, and successfully distills expression-related information exclusively in the  shape space. In Fig.~\ref{fig:MUG_1}-(b) we show that by interpolating the learned latent representations, we can generate meaningful shape interpolations that mimic facial expressions.  

In cases where data has a multi-modal distribution exhibiting multiple different canonical appearances, e.g., multi-class MNIST digit images, learning a single appearance is less meaningful and often challenging (Fig.~\ref{fig:mnist_all}-(b)). In such cases, utilizing class information (Sec.~\ref{section:class-aware}) significantly improves the quality of multi-modal appearance learning (Fig.~\ref{fig:mnist_all}-(d)). As the network learns to classify the images implicitly in its latent space, it learns to generate a single canonical appearance for each class. Misclassified data will be decoded into an incorrect class: the image at position (2,4) in Fig.~\ref{fig:mnist_all}-(c,d) is interpreted as a 6.

\begin{figure}[ht!]
 \centering
 \includegraphics[width=0.9\linewidth]{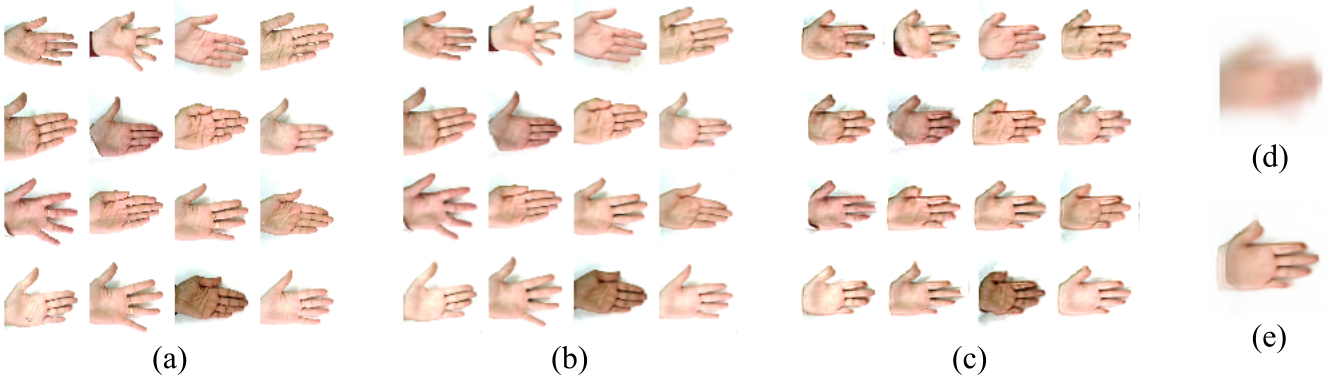}
 \caption{Unsupervised alignment on images of palms of left hands. (a) The input images; (b) reconstructed images; (c) texture images warped with the average of the decoded deformation; (d) the average input image; and (e) the average texture.}
 \label{fig:alpHandsNew}
  \vspace{-0.5cm}
\end{figure}

We now demonstrate the effectiveness of texture inference using our network on in-the-wild human faces. Using the MAFL face dataset, we show that our network is able to align the faces to a common texture space under various poses, illumination conditions, or facial expressions (Fig.~\ref{fig:intrinsicDecomp})-(d). The aligned textures retain the information of the input image such as lighting, gender, and facial hair, without a relevant supervision training signal. We further demonstrate the alignment on the 11k Hands dataset \cite{afifiHands}, where
we align palmar images of the left hand of several subjects \ref{fig:alpHandsNew}. This property of our network is especially useful for applications such as computer graphics, where  establishing correspondences (UV map) between a class of objects is important but usually difficult.

\subsection{Autoencoders vs. Deforming Autoencoders}

\begin{figure*}[!h]
    \centering
    \includegraphics[width=0.99\linewidth]{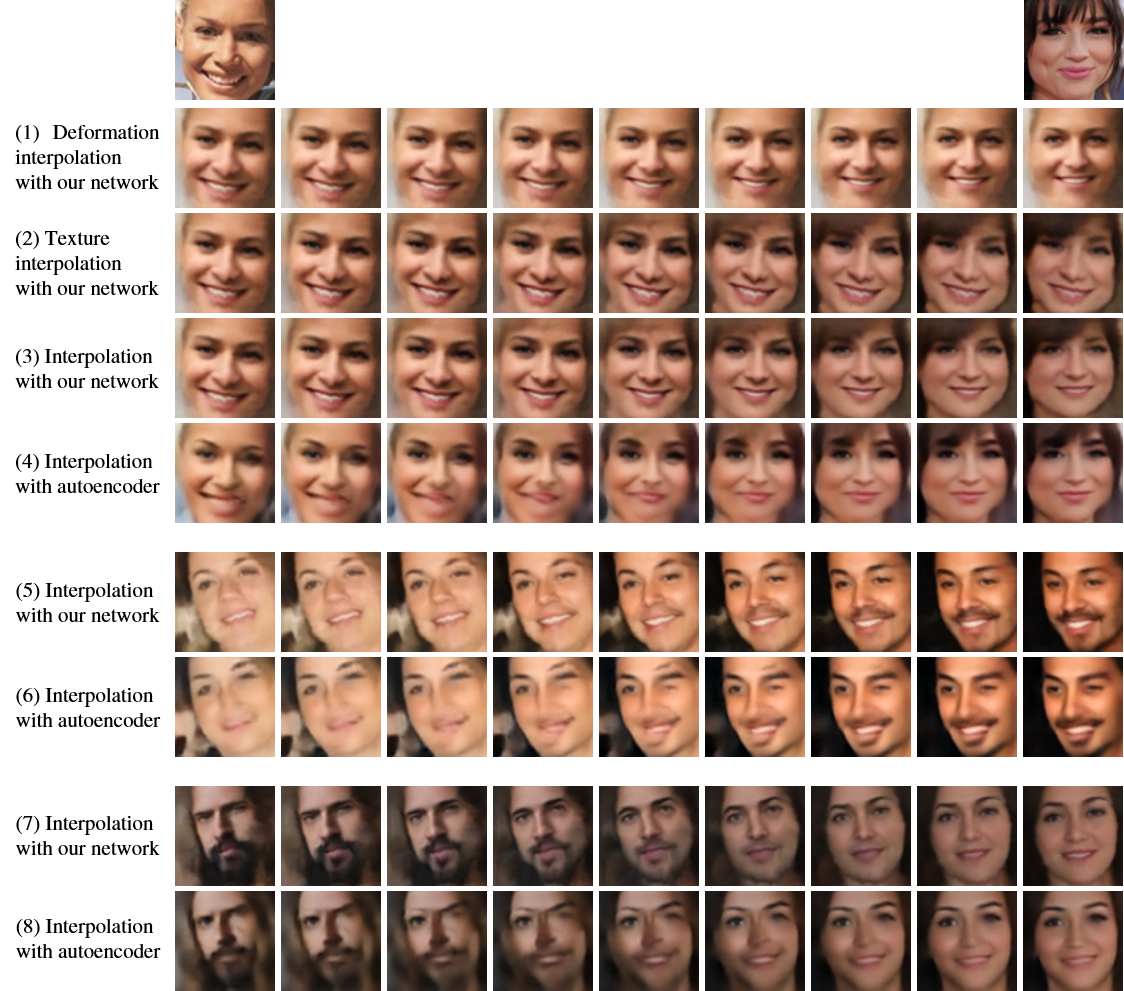}
    \caption{Latent representation interpolation: we embed a face image in the latent space provided by an encoder network trained on the MAFL dataset. Our network disentangles the texture and deformation in the respective parts of the latent representation vector,  allowing a meaningful interpolation between images. Interpolating the deformation-specific part of the latent representation changes the face shape and pose (1); interpolating the latent representation for texture will generate a pose-aligned texture transfer between the images (2); traversing both latent representations will generate smooth and sharp image deformations (3,5,7). In contrast, when using a standard auto-encoder (4,6,8) such an interpolation often yields artifacts. For more results, please see Figure~\ref{fig:supp_walk1},\ref{fig:supp_walk2} in Appendix.}
    \label{fig:traversal}
\end{figure*}

We show the ability of our network to learn meaningful deformation representations without supervision. We compare our disentangling network with a plain auto-encoder (Fig.~\ref{fig:traversal}). Contrary to our network which disentangles an image into a template texture and a deformation field, the auto-encoder is trained to encode all of the image in a single latent representation, i.e., the bottleneck.  

We train both networks in the MAFL faces-in-the-wild dataset. To evaluate the learned representation, we conduct manifold traversal (i.e., latent representation interpolation) between two randomly sampled face images: given a source face image $I^s$ and a target image $I^t$, we first compute their latent representations $Z$s. We use $Z_T(I^s)$ and $Z_S(I^s)$ to denote the latent representations in our network for $I^s$, and $Z_{ae}(I^s)$ for the latent representation learned by a plain autoencoder. We then conduct linear interpolation on $Z$, between $Z^s$ and $Z^t$: 
$Z^{\lambda} = \lambda Z^s + (1-\lambda) Z^t 
$.
We subsequently reconstruct the image $I^{\lambda}$ from $Z^{\lambda}$ using the corresponding decoder(s), as shown in ~\reffig{fig:traversal}.

By traversing the learned deformation representation only, we can change the shape and pose of a face while maintaining its texture (\reffig{fig:traversal}-(1)); interpolating the texture representation results in pose-aligned texture transfer (\reffig{fig:traversal}-(2)); traversing on both representations will generate a smooth deformation from one image to another (\reffig{fig:traversal}-(3,5,7)). Compared to the interpolation using the autoencoder (\reffig{fig:traversal}-(4,6,8)), which often exhibits artifacts, our traversal stays on the semantic manifold of faces and generates sharp facial features.

\subsection{Intrinsic Deforming Autoencoders}
\begin{figure}[ht!]
    \centering
    \includegraphics[width=0.99\linewidth]{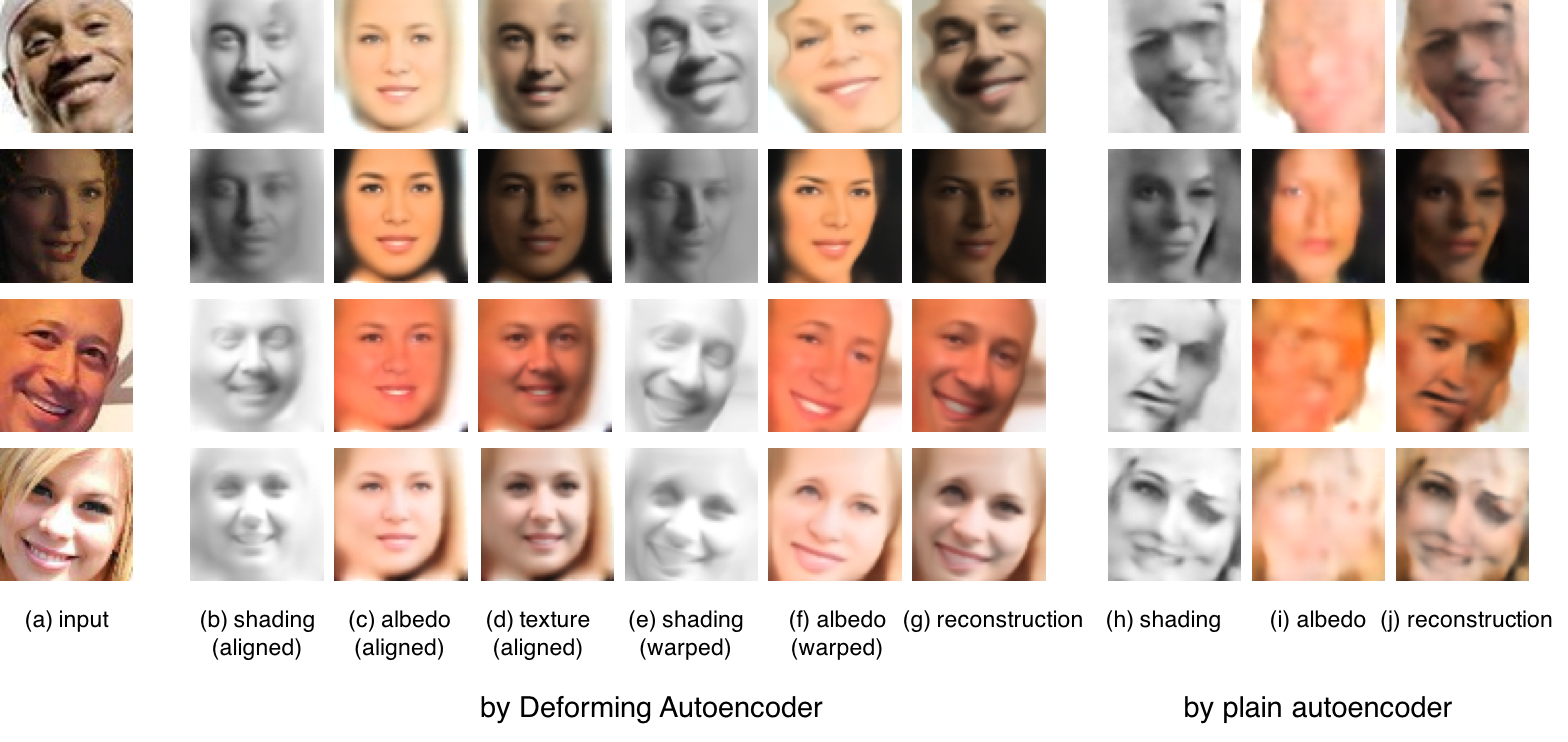}
    \caption{Unsupervised intrinsic decompostion with Deforming Autoencoders (Intrinsic-DAE). Thanks to the ``automatic dense aligment'' property of DAE, shading and albedo are faithfully separated (e,f) by the intrinsic decomposition loss. Shading (b) and albedo (c) are learned in an unsupervised manner in the densely aligned canonical space. With the  deformation field also learned  without supervision, we can recover the intrinsic image components for the original shape and viewpoint (e,f). Without dense alignment, the intrinsic decomposition loss fails to decompose shading and albedo (h,i,j).}
    \label{fig:intrinsicDecomp}
\end{figure}

Having demonstrated the disentanglement abilities of Deforming Autoencoders, we now  explore the disentanglement capabilities of Intrinsic-DAE described in Sec. \ref{sec:idae}. 
Using only the $E_{\text{DA}}$ and regularization losses, the Intrinsic-DAE is able to generate convincing shading and albedo estimates without direct supervision (Fig.~\ref{fig:intrinsicDecomp}-(b) to (g)).
Without the ``learning-to-align'' property, a baseline autoencoder structure with an intrinsic decomposition design (Fig.~\ref{fig:daeIntrinsicStructure}-(b)) cannot decompose the image into plausible shading and albedo components (Fig.~\ref{fig:intrinsicDecomp}-(h),(i),(j)).

\begin{figure}[ht!]
    \centering
    \includegraphics[width=0.92\linewidth]{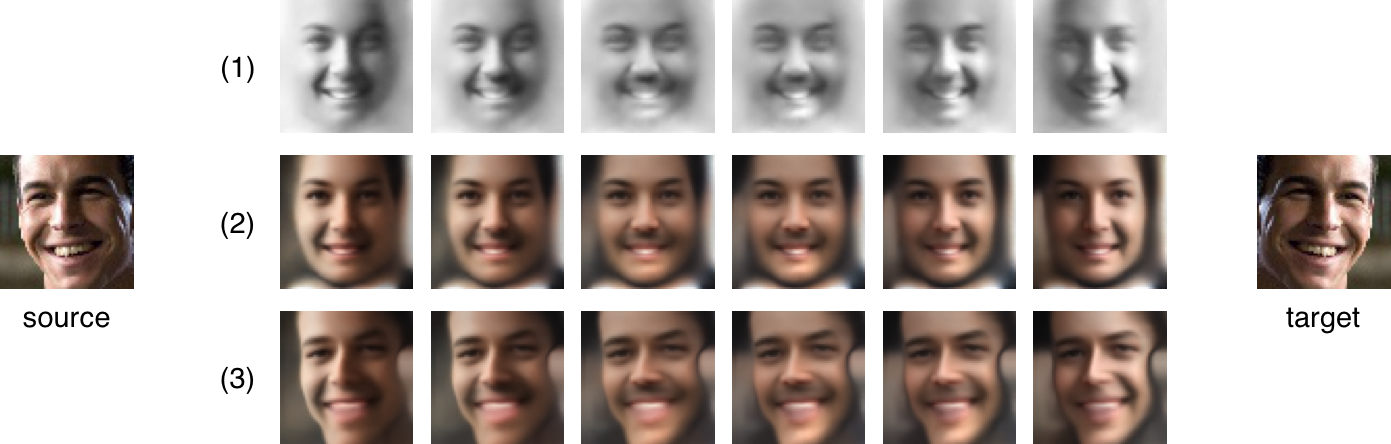}
    \caption{Lighting interpolation with Intrinsic-DAE. With latent representations learned in an unsupervised manner for shading, albedo, and defomation, the DAE allows us to simulate smooth transitions of the lighting direction. In this example, we interpolate the latent representation of the shading from source (lit from the left) to target (mirrored source, hence lit from the right). The network generates smooth lighting transitions, without explicitly learning geometry, as shown in shading (1) and texture (2). Together with the  learned deformation of the source image, DAE enables the relighting of the face in its original pose (3).}
    \label{fig:lightingInterpolation}
\end{figure}

In addition, we show that by manipulating the learned latent representation of $S$, Intrinsic-DAE allows us to simulate illumination effects for face images, such as interpolating lighting directions (Fig.~\ref{fig:lightingInterpolation}).

\begin{figure}[!h]
    \centering
    \includegraphics[width=0.85\linewidth]{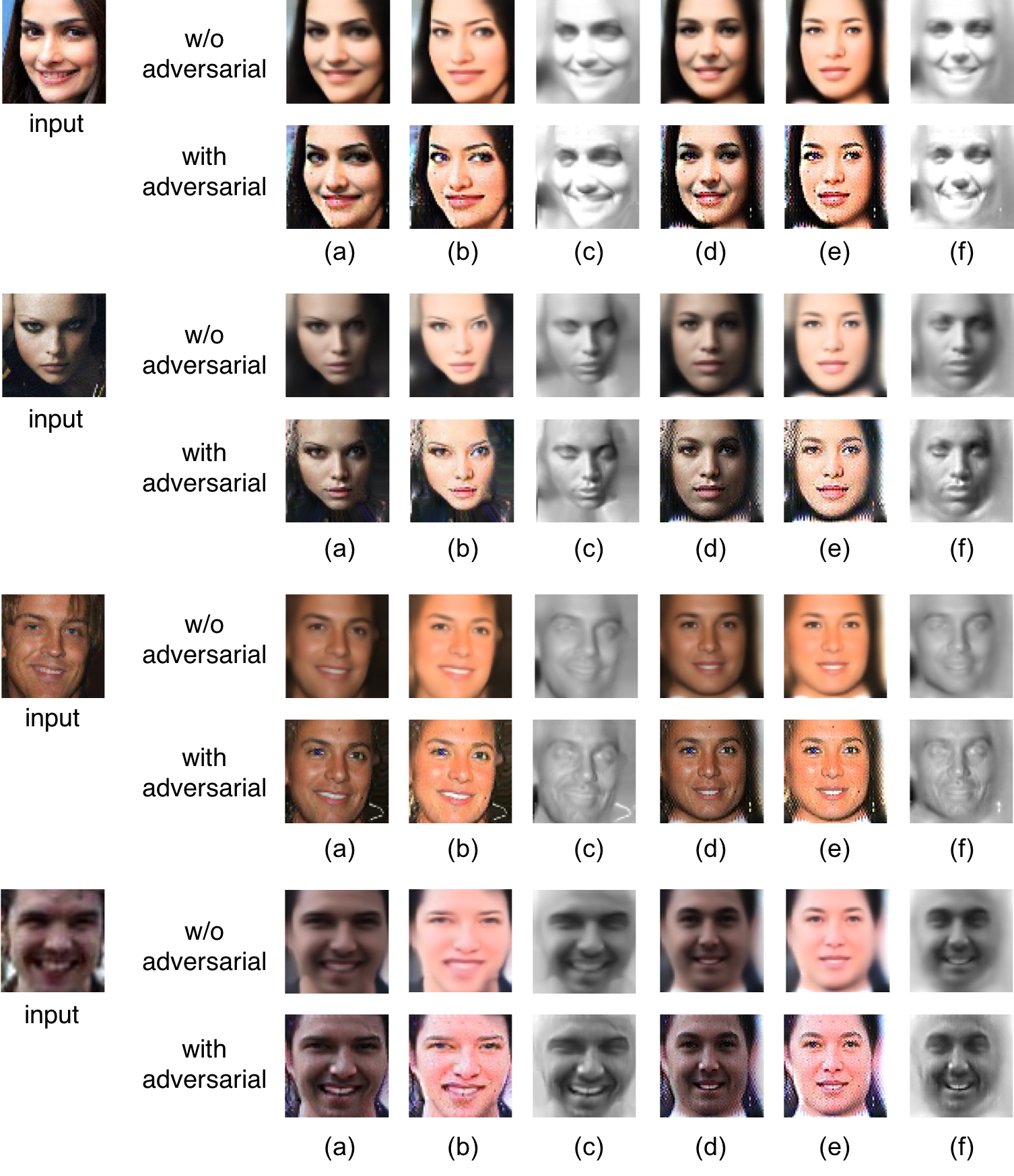}
    \caption{Intrinsic-DAE with an adversarial loss: (a/d) reconstruction (b/e) albedo, (c/f) shading, in image and template coordinates, respectively. 
    Applying an adversarial loss to the final output results improves the visual quality of the image reconstruction (a) of Intrinsic-DAE, while the deformation, albedo, and shading can still be successfully disentangled.}
    \label{fig:GAN}
\end{figure}

Training with $L2$ reconstruction losses, autoencoder-like architectures are prone to generating smooth images which lack visual realism (Fig.~\ref{fig:intrinsicDecomp}). Inspired by the success of generative adversarial networks (GANs)~\cite{goodfellow2014generative}, we follow previous work~\cite{ShuYHSSS17} where an adversarial loss is adopted to generate visually realistic images: we train the Intrinsic-DAE with an extra adversarial loss term $E_{\text{Adversarial}}$ applied on the final output. The loss function becomes:
\begin{equation}
    E_{\text{Intinsic-DAE}} = E_{\text{Reconstruction}} + E_{\text{Warp}} + \lambda_4 E_{\text{Adversarial}}.\textbf{}
\end{equation}
In practice, we apply a PatchGAN~\cite{li2016precomputed,pix2pix2016} as the discriminator and set $\lambda_4 = 0.1$. We found that the adversarial loss improves the visual sharpness of the reconstruction while the deformation, shading are still successfully disentangled (Fig. ~\ref{fig:GAN}).

\subsection{Unsupervised alignment evaluation}

\begin{figure}[!h]
    \centering
    \includegraphics[width=0.9\linewidth]{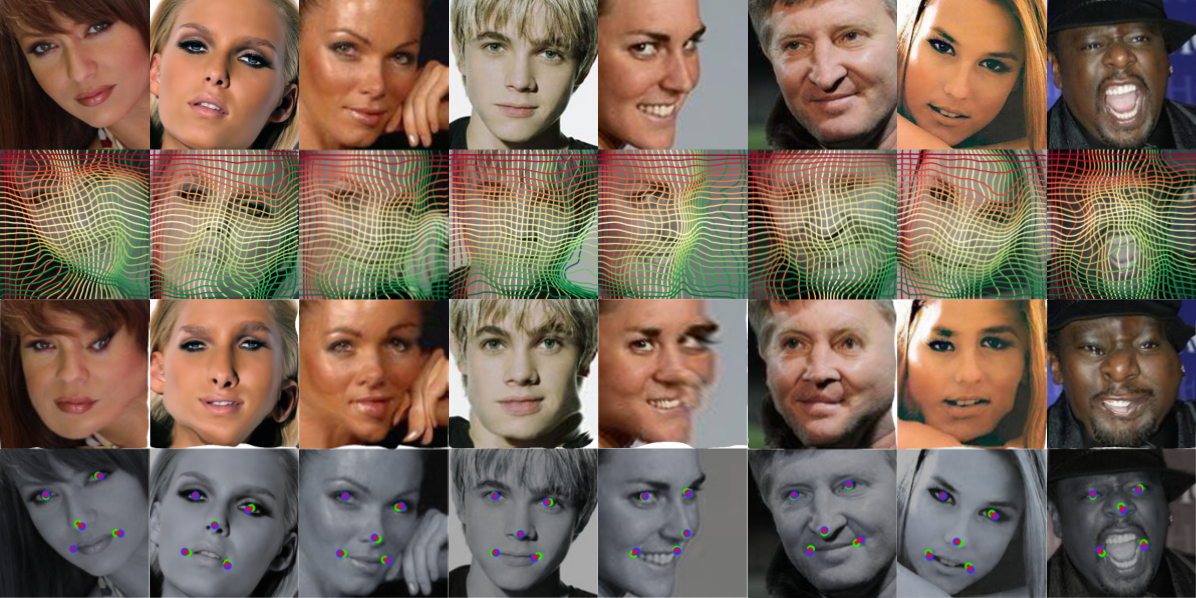}
    \caption{\emph{1st row}: Sample images from the MAFL test set; \emph{2nd row}:
    Estimated deformation grid; and \emph{3rd row}: Image reverse-transformed to texture space \emph{4th row}: semantic landmark locations (green: ground truth landmark locations, 
    blue: estimated landmark locations,
    red: error lines).}
    \label{fig:landmarksAndKeypoints}
\end{figure}
Having qualitatively analyzed the disentanglement capabilities of our networks, we now turn to quantifying their performance on the task of unsupervised image alignment. 
We report the performance of our face DAE's alignment on landmark detection on face images, specifically, the eyes, the nose, and corners of the mouth. 
We report performance on the MAFL dataset, which contains manually annotated landmark locations for 19,000 training and 1,000 test images.
In our experiments, we use a model trained on the CelebA dataset without any form of supervision to estimate deformation fields on the MAFL training set. Following the evaluation protocol of the work that we directly compare to \cite{ThewlisBV17a}, we train a landmark regressor post-hoc on these deformation fields using the provided annotations. We use landmark locations from the MAFL training set  as training data for this regressor, but do not pass gradients to the Deforming Autoencoder, which thereby remains fixed to the model learned without supervision. The regressor is a  2-layer fully-connected neural network. Its inputs are flattened deformation
  fields (vectors of size $64 \times 64 \times 2$), which are provided as input to a 100-dimensional hidden layer, followed by a ReLU and a 10-D output layer to predict the spatial coordinates ($(x, y)$) for five landmarks corresponding to the eyes, nose, and mouth corner landmarks. We use L1 loss as the objective function for this regression task.
  
In testing, we predict landmark locations using the trained regressor and the deformation fields on the MAFL test set. In Table 1 we report the mean error in landmark localization as a percentage of the inter-ocular distance. 
As the deformation field determines the alignment in the texture space, it serves as an effective mapping between landmark locations
on the aligned texture and those on the original, unaligned faces. Hence, the mean error we report directly quantifies the quality of
the (unsupervised) face alignment.  
  
  \mycomment{
  \begin{figure}[ht!]
 \centering
 \includegraphics[width=0.9\linewidth]{figures/eigenfaces_MAFLtrain.png}
 \caption{(a): First 100 eigen vectors computed from the set of textures decoded by the network.
    The set of faces used for this is the MAFL training set; (b) first 100 eigen vectors
    for the $X$-deformation; and (c) first 100 eigen vectors for the $Y$-deformation.}
  \label{fig:eigenfacesMAFL}
\end{figure}}

\begin{table}[ht!]
 \centering
 \begin{tabular}{c | c | c | c | c }
      $A$, MAFL & $I$, MAFL & $A+I$, MAFL & $A+I$, CelebA & $A+I$, CelebA, with Regressor  \\
      \hline
      14.13 & 9.89 & 8.50 & 7.54 & 5.96\\
      \hline
 \end{tabular}
 \caption{Improvement in landmark localization errors on the MAFL test set as we add new types of deformation and new data. In the table, $A$ indicates a model which uses the affine transformation, $I$ indicates one with the integral transformation, whereas MAFL and CelebA denote which dataset the deforming autoencoder was trained on.  For columns 1 to 4, we manually annotate landmarks on the average texture image, while for column 5, we train a regressor on the deformation fields to predict them. In all experiments, each latent vector in the DAE is of size 32.}
 \label{tab:errDeforms}
 \vspace{-.5cm}
\end{table}

In Table 2 we compare with the results of the best current method for semi-supervised image registration \cite{ThewlisBV17a}. We observe that by better modeling of the deformation space we quickly bridge the gap in performance, even though we never explicitly trained to learn correspondences.

\begin{table}[ht!]
 \vspace{-0.2cm}
 \centering
 \begin{tabular}{c c c c c c | c c c | c | c}
   \multicolumn{6}{c|}{DAE} & \multicolumn{3}{c|}{Dense-DAE} & \multirow{2}{*}{TCDCN\cite{zhang2016learning}} & \multirow{2}{*}{Thewlis et al.\cite{ThewlisBV17a}} \\
  \cline{1-9}
  32-NR & 32-Res & 16 & 32 & 64 & 96 & 16 & 64 & 96 & & \\
  \hline
  10.24 & 9.93 & 5.71 & 5.96 & 5.70 & 6.46 & 6.85 & 5.50 & \textbf{5.45} & 7.95 & 5.83 \\
  \hline
 \end{tabular}
 \caption{Mean error on unsupervised landmark detection on the MAFL test set, expressed as a percentage of the inter-ocular distance: modeling non-rigid deformations clearly reduces error more than just modeling affine ones. DAE and Dense-DAE denote two flavours of the deforming autoencoder - with and without dense convolutional connections, respectively. Under DAE and Dense-DAE we specify the size of each latent vector in the deforming autoencoder. \emph{NR} signifies training without regularization on the estimated deformations, while \emph{Res} signifies training by estimating the residual deformation grid instead of the integral. Our results clearly outperform the self-supervised method of \cite{ThewlisBV17a} trained specifically for establishing correspondences.}
 \label{tab:evalMAFL}
 \vspace{-0.8cm}
\end{table}

\section{Conclusion and Future Work}

In this paper we have  developed deep autoencoders that can disentangle shape and appearance  in latent representation space. 
We have  shown that this method can be used for unsupervised groupwise image alignment. Our experiments with expression morphing in humans, image manipulation, such as shape and appearance interpolation,  as well as unsupervised landmark localization, show the generality of our approach. 
We have shown that bringing images in a canonical coordinate system allows for a more extensive form of image disentangling, facilitating the estimation of  decompositions into shape, albedo and shading without any form of supervision. We expect that this will lead in the future to a full-fledged disentanglement into normals, illumination, and 3D geometry.

\section{Acknowledgment}

This work was supported by a gift from Adobe, NSF grants CNS-1718014 and DMS 1737876, the Partner University Fund, and the SUNY2020 Infrastructure Transportation Security Center.

\bibliographystyle{splncs}
\bibliography{egbib}

\clearpage

\appendix
\section{Architectural Details}
\label{append:architecture}
\subsection{Convolutional Encoders and Decoders}
\label{sec:convnet}
In our experiments, where input images are of size $64 \times 64 \times \text{Nc}$ (Nc is 1 for MNIST and 3 for faces), we use identical architectures for convolutional encoders and decoders.

The encoder architecture is
\begin{verbatim}
Conv(32)-LeakyReLU-Conv(64)-BN-LeakyReLU-Conv(128)->
 ->BN-LeakyReLU-Conv(256)-BN-LeakyReLU-Conv(Nz)->
 ->Sigmoid;
\end{verbatim}
while the decoder architecture is
\begin{verbatim}
ConvT(256)-BN-ReLU-ConvT(128)-BN-ReLU-ConvT(64)->
 ->BN-ReLU-ConvT(32)-BN-ReLU-ConvT(32)-BN-ReLU-ConvT(Nc)->
 ->Threshold(0,1),
\end{verbatim}

where
\begin{itemize}
	\item \texttt{Conv(n)}: convolution layer with $n$ output feature map;
	\item \texttt{ConvT(n)}: transposed convolution (deconvolution) layer with $n$ output feature map;
	\item \texttt{BN}: batch normalization layer
	\item \texttt{Nz}: latent representation dimension
	\item \texttt{Nc}: number of output image channel
\end{itemize}

\subsection{DenseNet-style Encoders and Decoders}
\label{sec:densenet}
For DenseNet-style architectures, we employ dense convolutional connections. The architecture for the 
encoder is 
\begin{verbatim}
BN-ReLU-Conv(32)-DBE(32,6)-TBE(32,64,2)->
 ->DBE(64,12)-TBE(64,128,2)-DBE(128,24)-TBE(128,256,2)->
 ->DBE(256,16)-TBE(256,Nz,4)-Sigmoid;
\end{verbatim}
whereas the architecture for the decoder is
\begin{verbatim}
BN-Tanh-ConvT(256)-DBD(256,16)-TBD(256,128)->
 ->DBD(128,24)-TBD(128,64)-DBD(64,12)-TBD(64,32)->
 ->DBD(32,6)-TBD(32,32)-BN-Tanh-ConvT(Nc)-Threshold(0,1),
\end{verbatim}
where
\begin{itemize}
    \item \texttt{DBE(n,k)}: A dense encoder block with $k$ $3\times 3$ convolutions with $n$ channels. 
    \item \texttt{TBE(m,n,p)}: An encoder transition block of $1 \times 1$ convolutions with $m$ input channels and $n$ output channels. Also includes a max-pooling operation of size $p$.
    \item \texttt{DBD(n,k)}: A dense decoder block with $k$ $3 \times 3$ transposed convolution operations with $n$ channels. 
    \item \texttt{TBD(m,n)}: A decoder transition block of $4 \times 4$ convolutions, stride of $2$ and padding of $1$. It has $m$ input channels, and $n$ output channels.
\end{itemize}

We describe the tensor sizes for intermediate convolution operations in Tables \ref{tab:conv_architectures} and \ref{tab:dense_architectures}.

\begin{table}[ht!]
  \centering
  \begin{tabular}{|c|c|c|c|}
    \hline
    \multicolumn{2}{|c|}{Conv Encoder} & \multicolumn{2}{c|}{Conv Decoder} \\
    \hline
    Output Size & Operation & Output size & Operation \\
    \hline
    \hline
    $32 \times 32 \times 32$ & $4 \times 4$ Conv$(32)$ & $4 \times 4 \times 256$ & $4 \times 4$ ConvT$(256)$ \\
    \hline
    $16 \times 16 \times 64$ & $4 \times 4$ Conv$(64)$ & $4 \times 4 \times 128$ & $4 \times 4$ ConvT$(128)$ \\
    \hline
    $8 \times 8 \times 128$ & $4 \times 4$ Conv$(128)$ & $4 \times 4 \times 64$ &  $4 \times 4$ ConvT$(64)$ \\
    \hline
    $4 \times 4 \times 256$ & $4 \times 4$ Conv$(256)$ & $4 \times 4 \times 32$ &  $4 \times 4$ ConvT$(32)$ \\
    \hline
    \texttt{Nz}             & $4 \times 4$ Conv(\texttt{Nz}) & $4 \times 4 \times 32$ & $4 \times 4$ ConvT$(32)$ \\
    \hline
    & & $4 \times 4 \times $\texttt{Nc} & $4 \times 4$ ConvT(\texttt{Nc}) \\
    \hline
  \end{tabular}
  \caption{Tensor sizes for intermediate convolutional operations in the convolutional encoder and decoder architectures.
  The output shape denoted $h \times w \times C$, where $h$ and $w$ are height and width of the feature maps, respectively, 
  and $C$ is the number of channels.}
  \label{tab:conv_architectures}
\end{table}

\begin{table}[ht!]
  \centering
  \begin{tabular}{|c|c|c|c|}
    \hline
    \multicolumn{2}{|c|}{Dense Conv Encoder} & \multicolumn{2}{c|}{Dense Conv Decoder} \\
    \hline
    Output Size & Operation & Output size & Operation \\
    \hline
    \hline
    $32 \times 32 \times 32$ & $4 \times 4$ Conv$(32)$ & $4 \times 4 \times 256$ & $4 \times 4$ ConvT$(256)$ \\
    \hline
    $32 \times 32 \times 32$ & \texttt{DBE(32,6)} & $4 \times 4 \times 256$ & \texttt{DBD(256,16)}\\
    \hline
    $16 \times 16 \times 64$ & \texttt{TBE(32,64,2)} & $8 \times 8 \times 128$ &  \texttt{TBD(256,128)} \\
    \hline
    $16 \times 16 \times 64$ & \texttt{DBE(64,12)} & $8 \times 8 \times 128$ & \texttt{DBD(128,24)} \\
    \hline
    $8 \times 8 \times 128$ & \texttt{TBE(64,128,2)} & $16 \times 16 \times 64$ & \texttt{TBD(128,64)} \\
    \hline
    $8 \times 8 \times 128$ & \texttt{DBE(128,24)} & $16 \times 16 \times 64$ & \texttt{DBD(64,12)} \\
    \hline    
    $4 \times 4 \times 256$ & \texttt{TBE(128,256,2)} & $32 \times 32 \times 32$ & \texttt{TBD(64,32)} \\
    \hline    
    $4 \times 4 \times 256$ & \texttt{DBE(256,16)} & $32 \times 32 \times 32$ & \texttt{DBD(32,6)} \\
    \hline    
    \texttt{Nz}             & \texttt{TBE(256,\texttt{Nz},4)} & $64 \times 64 \times 32$ & \texttt{TBD(32,32)} \\
    \hline    
                          &                                & $64 \times 64 \times $\texttt{Nc} & $3 \times 3$ ConvT(\texttt{Nc}) \\
    \hline         
  \end{tabular}
  \caption{Tensor sizes for intermediate convolutional operations in the dense encoder and decoder architectures.
  The output shape denoted $h \times w \times C$, where $h$ and $w$ are height and width of the feature maps, respectively, 
  and $C$ is the number of channels.}
  \label{tab:dense_architectures}
\end{table}





\section{Ablation Study}
\label{append:ablation}
\subsection{Dimension of $Z_T$}

\begin{figure}[ht]
	\begin{center}
		\begin{tabular}{c@{\hspace{0.1in}}c@{\hspace{0.1in}}c@{\hspace{0.1in}}c@{\hspace{0.1in}}c}
		(a) input & 
		\includegraphics[height=0.8in]{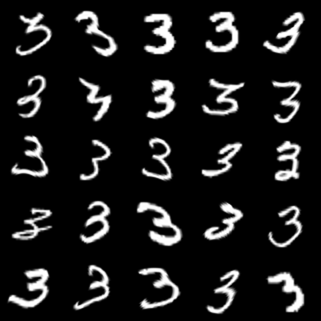} & & &  \\~\\~\\
		
		(b) 0-D $Z_T$ &
		\includegraphics[height=0.9in]{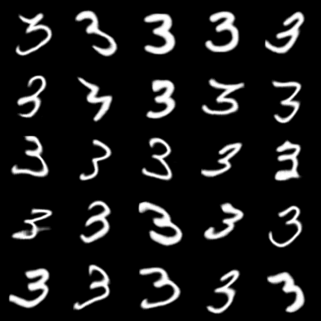} &
		\includegraphics[height=0.9in]{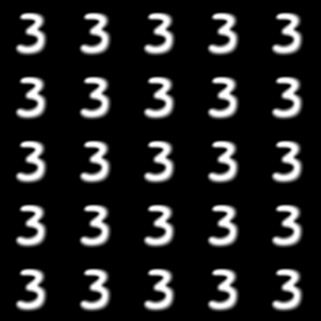} &
		\includegraphics[height=0.9in]{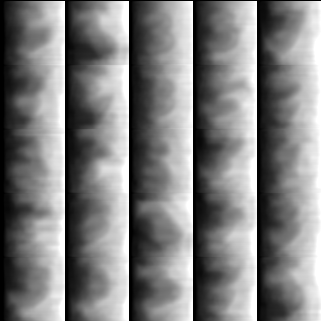} &
		\includegraphics[height=0.9in]{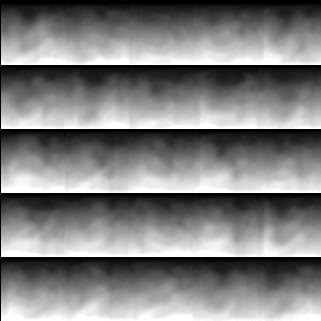} \\
		
		(c) 1-D $Z_T$ &
		\includegraphics[height=0.9in]{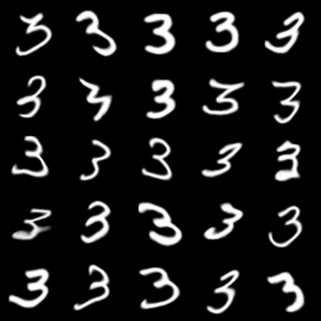} &
		\includegraphics[height=0.9in]{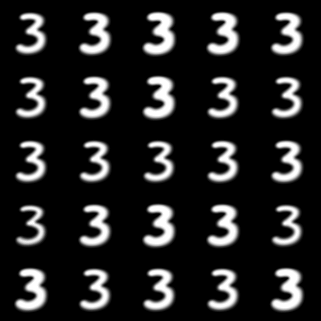} &
		\includegraphics[height=0.9in]{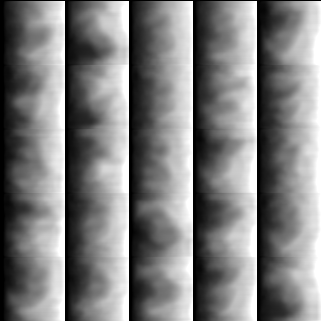} &
		\includegraphics[height=0.9in]{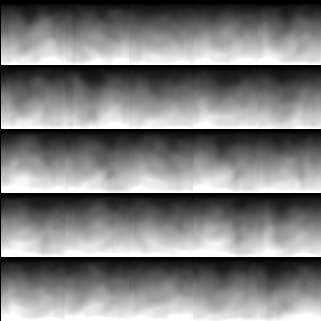} \\
		
		(d) 4-D $Z_T$ &
		\includegraphics[height=0.9in]{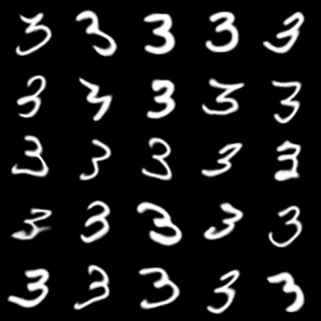} &
		\includegraphics[height=0.9in]{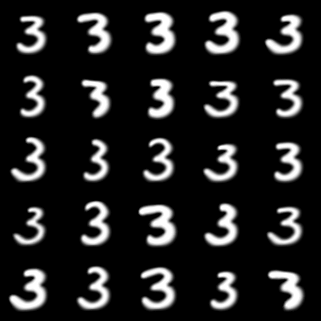} &
		\includegraphics[height=0.9in]{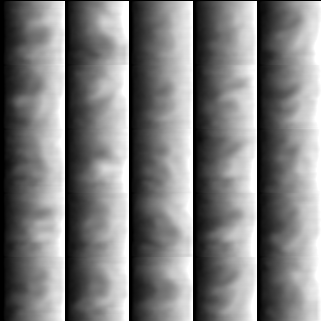} &
		\includegraphics[height=0.9in]{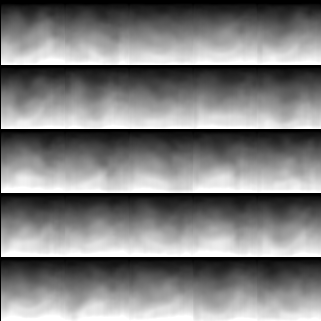} \\
		
		(e) 8-D $Z_T$ &
		\includegraphics[height=0.9in]{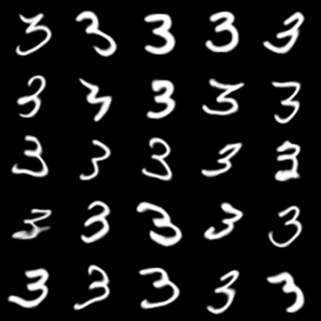} &
		\includegraphics[height=0.9in]{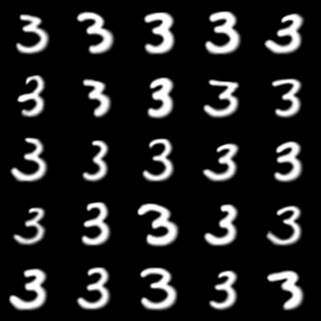} &
		\includegraphics[height=0.9in]{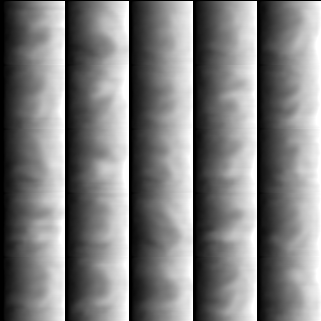} &
		\includegraphics[height=0.9in]{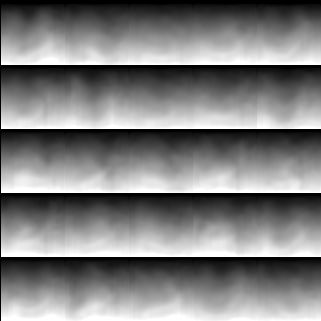} \\
		
		(f) 16-D $Z_T$ &
		\includegraphics[height=0.9in]{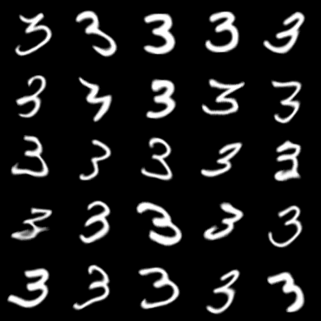} &
		\includegraphics[height=0.9in]{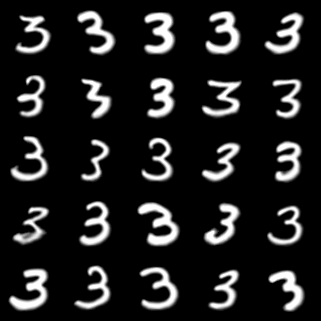} &
		\includegraphics[height=0.9in]{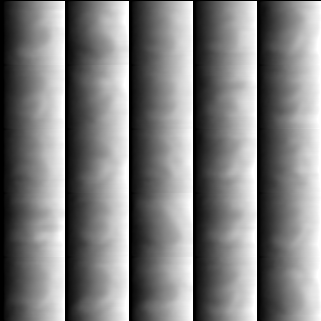} &
		\includegraphics[height=0.9in]{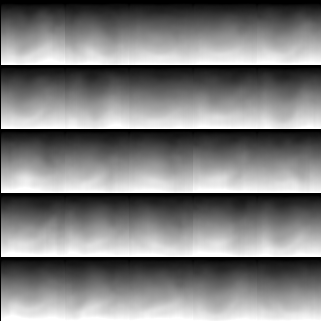} \\~\\
		
		&
		reconstruction & texture & warping (x) & warping (y)
			
		\end{tabular}
	\end{center}
	\caption{Effect of varying the dimensionality of the latent vector for the texture encoding, $Z_T$: The dimension of $Z_T$ is 0 for (b), 1 for (c), 4 for (d), 8 for (e), 16 for (f). $Z_W$ is fixed to 128. When $Z_T$ is 0-Dimensional, the texture decoder is forced to generate an identical texture for every image (b). When we increase the dimension of $Z_T$ to 1, the texture decoder learns to align the pose (c) with varying stroke width. When further increasing the dimension of $Z_T$, the network learns a more diverse texture map for each image (d, e, f). }
\label{fig:supp_dimension_mnist}
\end{figure}

\begin{figure}[ht]
	\begin{center}
		\begin{tabular}{c@{\hspace{0.15in}}c@{\hspace{0.02in}}c@{\hspace{0.02in}}c@{\hspace{0.02in}}c@{\hspace{0.02in}}c@{\hspace{0.02in}}c@{\hspace{0.02in}}c@{\hspace{0.02in}}c@{\hspace{0.02in}}c@{\hspace{0.02in}}c}
		Input &
		\includegraphics[height=0.35in]{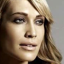} &
		\includegraphics[height=0.35in]{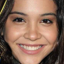} &
		\includegraphics[height=0.35in]{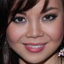} &
		\includegraphics[height=0.35in]{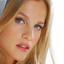} &
		\includegraphics[height=0.35in]{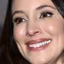} &
		\includegraphics[height=0.35in]{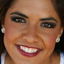} &
		\includegraphics[height=0.35in]{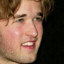} &
		\includegraphics[height=0.35in]{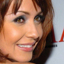} &
		\includegraphics[height=0.35in]{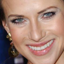} &
		\includegraphics[height=0.35in]{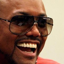} \\~\\
		
		Reconstruction, 0-D &
		\includegraphics[height=0.35in]{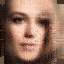} &
		\includegraphics[height=0.35in]{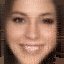} &
		\includegraphics[height=0.35in]{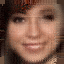} &
		\includegraphics[height=0.35in]{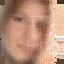} &
		\includegraphics[height=0.35in]{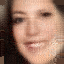} &
		\includegraphics[height=0.35in]{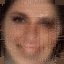} &
		\includegraphics[height=0.35in]{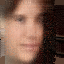} &
		\includegraphics[height=0.35in]{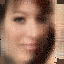} &
		\includegraphics[height=0.35in]{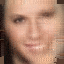} &
		\includegraphics[height=0.35in]{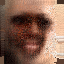} \\
		
		Texture, 0D &
		\includegraphics[height=0.35in]{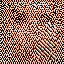} &
		\includegraphics[height=0.35in]{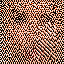} &
		\includegraphics[height=0.35in]{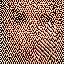} &
		\includegraphics[height=0.35in]{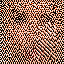} &
		\includegraphics[height=0.35in]{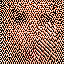} &
		\includegraphics[height=0.35in]{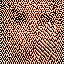} &
		\includegraphics[height=0.35in]{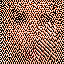} &
		\includegraphics[height=0.35in]{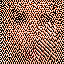} &
		\includegraphics[height=0.35in]{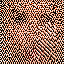} &
		\includegraphics[height=0.35in]{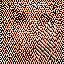} \\~\\

        Reconstruction, 2D &
		\includegraphics[height=0.35in]{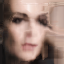} &
		\includegraphics[height=0.35in]{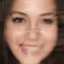} &
		\includegraphics[height=0.35in]{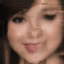} &
		\includegraphics[height=0.35in]{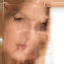} &
		\includegraphics[height=0.35in]{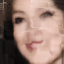} &
		\includegraphics[height=0.35in]{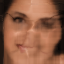} &
		\includegraphics[height=0.35in]{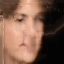} &
		\includegraphics[height=0.35in]{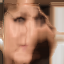} &
		\includegraphics[height=0.35in]{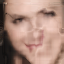} &
		\includegraphics[height=0.35in]{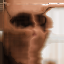} \\
		
		Texture, 2D &
		\includegraphics[height=0.35in]{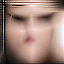} &
		\includegraphics[height=0.35in]{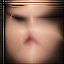} &
		\includegraphics[height=0.35in]{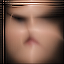} &
		\includegraphics[height=0.35in]{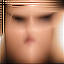} &
		\includegraphics[height=0.35in]{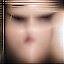} &
		\includegraphics[height=0.35in]{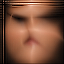} &
		\includegraphics[height=0.35in]{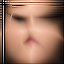} &
		\includegraphics[height=0.35in]{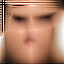} &
		\includegraphics[height=0.35in]{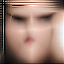} &
		\includegraphics[height=0.35in]{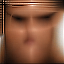} \\~\\
		
        Reconstruction, 4D &
		\includegraphics[height=0.35in]{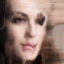} &
		\includegraphics[height=0.35in]{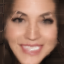} &
		\includegraphics[height=0.35in]{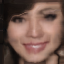} &
		\includegraphics[height=0.35in]{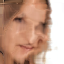} &
		\includegraphics[height=0.35in]{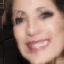} &
		\includegraphics[height=0.35in]{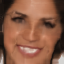} &
		\includegraphics[height=0.35in]{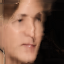} &
		\includegraphics[height=0.35in]{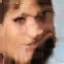} &
		\includegraphics[height=0.35in]{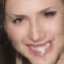} &
		\includegraphics[height=0.35in]{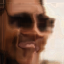} \\
		
		Texture, 4D &
		\includegraphics[height=0.35in]{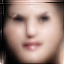} &
		\includegraphics[height=0.35in]{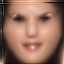} &
		\includegraphics[height=0.35in]{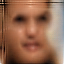} &
		\includegraphics[height=0.35in]{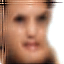} &
		\includegraphics[height=0.35in]{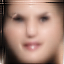} &
		\includegraphics[height=0.35in]{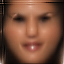} &
		\includegraphics[height=0.35in]{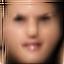} &
		\includegraphics[height=0.35in]{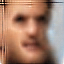} &
		\includegraphics[height=0.35in]{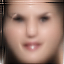} &
		\includegraphics[height=0.35in]{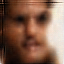} \\~\\

        Reconstruction, 16D &
		\includegraphics[height=0.35in]{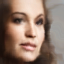} &
		\includegraphics[height=0.35in]{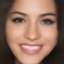} &
		\includegraphics[height=0.35in]{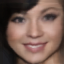} &
		\includegraphics[height=0.35in]{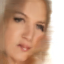} &
		\includegraphics[height=0.35in]{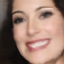} &
		\includegraphics[height=0.35in]{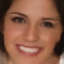} &
		\includegraphics[height=0.35in]{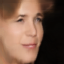} &
		\includegraphics[height=0.35in]{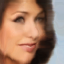} &
		\includegraphics[height=0.35in]{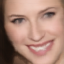} &
		\includegraphics[height=0.35in]{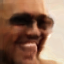} \\
		
		Texture, 16D &
		\includegraphics[height=0.35in]{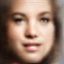} &
		\includegraphics[height=0.35in]{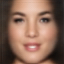} &
		\includegraphics[height=0.35in]{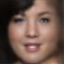} &
		\includegraphics[height=0.35in]{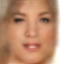} &
		\includegraphics[height=0.35in]{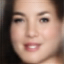} &
		\includegraphics[height=0.35in]{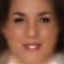} &
		\includegraphics[height=0.35in]{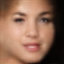} &
		\includegraphics[height=0.35in]{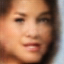} &
		\includegraphics[height=0.35in]{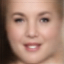} &
		\includegraphics[height=0.35in]{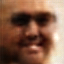} \\~\\  	

        Reconstruction, 32D &
		\includegraphics[height=0.35in]{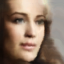} &
		\includegraphics[height=0.35in]{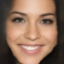} &
		\includegraphics[height=0.35in]{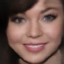} &
		\includegraphics[height=0.35in]{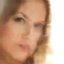} &
		\includegraphics[height=0.35in]{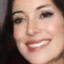} &
		\includegraphics[height=0.35in]{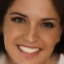} &
		\includegraphics[height=0.35in]{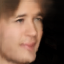} &
		\includegraphics[height=0.35in]{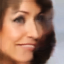} &
		\includegraphics[height=0.35in]{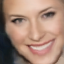} &
		\includegraphics[height=0.35in]{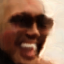} \\
		
		Texture, 32D &
		\includegraphics[height=0.35in]{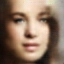} &
		\includegraphics[height=0.35in]{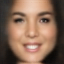} &
		\includegraphics[height=0.35in]{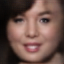} &
		\includegraphics[height=0.35in]{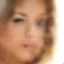} &
		\includegraphics[height=0.35in]{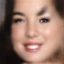} &
		\includegraphics[height=0.35in]{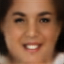} &
		\includegraphics[height=0.35in]{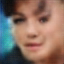} &
		\includegraphics[height=0.35in]{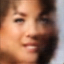} &
		\includegraphics[height=0.35in]{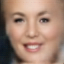} &
		\includegraphics[height=0.35in]{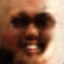} \\~\\

        Reconstruction, 128D &
		\includegraphics[height=0.35in]{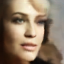} &
		\includegraphics[height=0.35in]{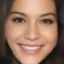} &
		\includegraphics[height=0.35in]{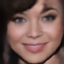} &
		\includegraphics[height=0.35in]{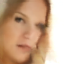} &
		\includegraphics[height=0.35in]{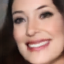} &
		\includegraphics[height=0.35in]{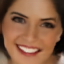} &
		\includegraphics[height=0.35in]{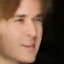} &
		\includegraphics[height=0.35in]{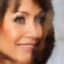} &
		\includegraphics[height=0.35in]{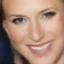} &
		\includegraphics[height=0.35in]{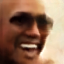} \\
		
		Texture, 128D &
		\includegraphics[height=0.35in]{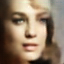} &
		\includegraphics[height=0.35in]{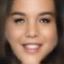} &
		\includegraphics[height=0.35in]{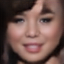} &
		\includegraphics[height=0.35in]{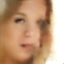} &
		\includegraphics[height=0.35in]{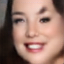} &
		\includegraphics[height=0.35in]{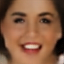} &
		\includegraphics[height=0.35in]{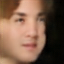} &
		\includegraphics[height=0.35in]{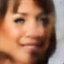} &
		\includegraphics[height=0.35in]{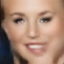} &
		\includegraphics[height=0.35in]{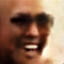} \\    					
		\end{tabular}
	\end{center}
	\caption{Effect of varying the dimensionality of the latent vector for the texture envoding,  $Z_T$, on the MAFL face dataset; $Z_W$ is fixed to 128. The problem is ill-posed and affords many solutions; if $Z_T$ is set to be 0D dimension, the texture becomes a ``bag of colored pixels'' which, when deformed (at will) can reconstruct an image. Increasing the dimension of $Z_T$ (4-32D) lets the network generates aligned texture maps  and more exact appearance; further increasing $Z_T$ (128-D) reduces the alignment effect. }
	\label{fig:supp_dimension_face}
\end{figure}

In this section, we show experimental results on single deformed MNIST images of the digit 3 (Figure \ref{fig:supp_dimension_mnist}) as well as in-the-wild faces (without masking) from the MAFL dataset (Figure \ref{fig:supp_dimension_face}) to demonstrate the effect of varying the dimension of $Z_T$.

\FloatBarrier
  
\subsection{Methods for deformation modeling}

In this section, we demonstrate the effect of using different warping modules.

\begin{figure}[ht]
		\begin{center}
		\begin{tabular}{c@{\hspace{0.1in}}c@{\hspace{0.02in}}c@{\hspace{0.02in}}c@{\hspace{0.02in}}c@{\hspace{0.02in}}c@{\hspace{0.02in}}c@{\hspace{0.02in}}c@{\hspace{0.02in}}c@{\hspace{0.02in}}c}
		
		(a) Image &
		\includegraphics[height=0.38in]{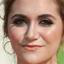} &
		\includegraphics[height=0.38in]{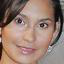} &
		\includegraphics[height=0.38in]{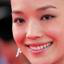} &
		\includegraphics[height=0.38in]{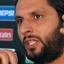} &
		\includegraphics[height=0.38in]{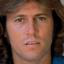} &
		\includegraphics[height=0.38in]{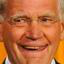} &
		\includegraphics[height=0.38in]{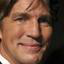} &
		\includegraphics[height=0.38in]{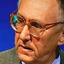} &
		\includegraphics[height=0.38in]{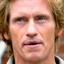} \\
		
		(b)-1 reconstruction&
        \includegraphics[height=0.38in]{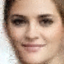} &
		\includegraphics[height=0.38in]{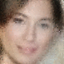} &
		\includegraphics[height=0.38in]{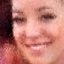} &
		\includegraphics[height=0.38in]{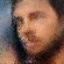} &
		\includegraphics[height=0.38in]{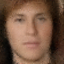} &
		\includegraphics[height=0.38in]{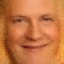} &
		\includegraphics[height=0.38in]{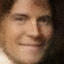} &
		\includegraphics[height=0.38in]{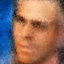} &
		\includegraphics[height=0.38in]{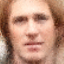} \\
		
		(b)-2 texture&
        \includegraphics[height=0.38in]{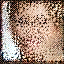} &
		\includegraphics[height=0.38in]{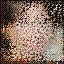} &
		\includegraphics[height=0.38in]{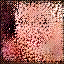} &
		\includegraphics[height=0.38in]{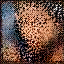} &
		\includegraphics[height=0.38in]{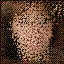} &
		\includegraphics[height=0.38in]{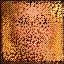} &
		\includegraphics[height=0.38in]{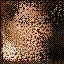} &
		\includegraphics[height=0.38in]{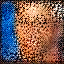} &
		\includegraphics[height=0.38in]{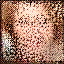} \\
		
		(b)-3 warping (x)&
        \includegraphics[height=0.38in]{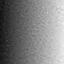} &
		\includegraphics[height=0.38in]{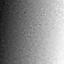} &
		\includegraphics[height=0.38in]{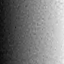} &
		\includegraphics[height=0.38in]{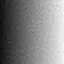} &
		\includegraphics[height=0.38in]{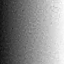} &
		\includegraphics[height=0.38in]{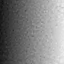} &
		\includegraphics[height=0.38in]{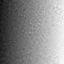} &
		\includegraphics[height=0.38in]{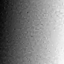} &
		\includegraphics[height=0.38in]{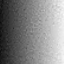} \\
		
		(c)-1 reconstruction&
		\includegraphics[height=0.38in]{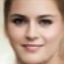} &
		\includegraphics[height=0.38in]{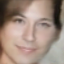} &
		\includegraphics[height=0.38in]{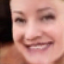} &
		\includegraphics[height=0.38in]{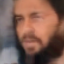} &
		\includegraphics[height=0.38in]{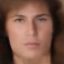} &
		\includegraphics[height=0.38in]{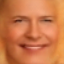} &
		\includegraphics[height=0.38in]{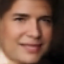} &
		\includegraphics[height=0.38in]{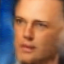} &
		\includegraphics[height=0.38in]{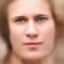} \\
		
		(c)-2 texture&
		\includegraphics[height=0.38in]{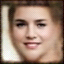} &
		\includegraphics[height=0.38in]{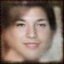} &
		\includegraphics[height=0.38in]{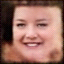} &
		\includegraphics[height=0.38in]{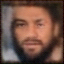} &
		\includegraphics[height=0.38in]{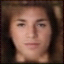} &
		\includegraphics[height=0.38in]{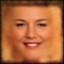} &
		\includegraphics[height=0.38in]{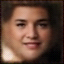} &
		\includegraphics[height=0.38in]{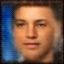} &
		\includegraphics[height=0.38in]{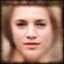} \\
		
		(c)-3 warping (x)&
		\includegraphics[height=0.38in]{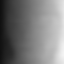} &
		\includegraphics[height=0.38in]{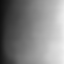} &
		\includegraphics[height=0.38in]{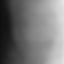} &
		\includegraphics[height=0.38in]{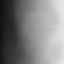} &
		\includegraphics[height=0.38in]{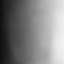} &
		\includegraphics[height=0.38in]{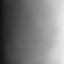} &
		\includegraphics[height=0.38in]{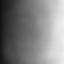} &
		\includegraphics[height=0.38in]{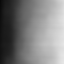} &
		\includegraphics[height=0.38in]{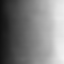} \\~\\~\\

		\end{tabular}
		\end{center}
\caption{Comparison between using our proposed \emph{affine + integral} warping modules (c) and using a warping field directly predicted from a convolutional decoder (b) for non-rigid deformation modeling. Our non-rigid deformation modeling generates better reconstructions and visually plausible texture maps. }
\label{fig:supp_integral_vs_direct}
\end{figure}

We first show additional comparisons between using our proposed \emph{affine + integral} warping and a non-rigid warping field directly output from a convolutional decoder for non-rigid deformation modeling (Figure \ref{fig:supp_integral_vs_direct}).

\begin{figure}[ht]
	\begin{center}
		\begin{tabular}{c@{\hspace{0.02in}}c@{\hspace{0.02in}}c@{\hspace{0.02in}}c@{\hspace{0.02in}}c@{\hspace{0.02in}}c@{\hspace{0.02in}}c@{\hspace{0.02in}}c@{\hspace{0.02in}}c@{\hspace{0.02in}}c}
            
		(a) Image &
		\includegraphics[height=0.40in]{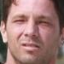} &
		\includegraphics[height=0.40in]{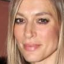} &
		\includegraphics[height=0.40in]{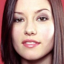} &
		\includegraphics[height=0.40in]{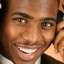} &
		\includegraphics[height=0.40in]{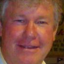} &
		\includegraphics[height=0.40in]{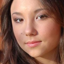} &
		\includegraphics[height=0.40in]{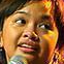} &
		\includegraphics[height=0.40in]{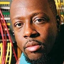} \\~\\~\\
		
		\textbf{Affine}:  & & & & & & & & \\~\\
		(b)-1 reconstruction &
		\includegraphics[height=0.40in]{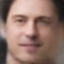} &
		\includegraphics[height=0.40in]{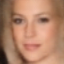} &
		\includegraphics[height=0.40in]{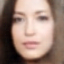} &
		\includegraphics[height=0.40in]{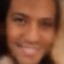} &
		\includegraphics[height=0.40in]{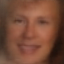} &
		\includegraphics[height=0.40in]{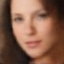} &
		\includegraphics[height=0.40in]{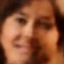} &
		\includegraphics[height=0.40in]{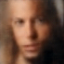} \\    
		(b)-2 texture &
		\includegraphics[height=0.40in]{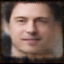} &
		\includegraphics[height=0.40in]{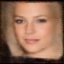} &
		\includegraphics[height=0.40in]{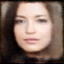} &
		\includegraphics[height=0.40in]{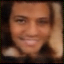} &
		\includegraphics[height=0.40in]{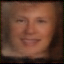} &
		\includegraphics[height=0.40in]{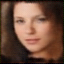} &
		\includegraphics[height=0.40in]{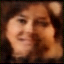} &
		\includegraphics[height=0.40in]{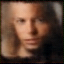} \\~\\~\\
		
		\textbf{Integral}: & & & & & & & & \\~\\
		(c)-1 reconstruction &
		\includegraphics[height=0.40in]{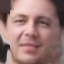} &
		\includegraphics[height=0.40in]{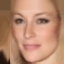} &
		\includegraphics[height=0.40in]{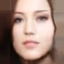} &
		\includegraphics[height=0.40in]{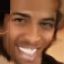} &
		\includegraphics[height=0.40in]{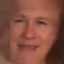} &
		\includegraphics[height=0.40in]{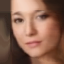} &
		\includegraphics[height=0.40in]{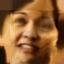} &
		\includegraphics[height=0.40in]{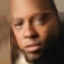} \\    
		(c)-2 texture &
		\includegraphics[height=0.40in]{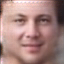} &
		\includegraphics[height=0.40in]{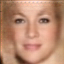} &
		\includegraphics[height=0.40in]{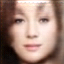} &
		\includegraphics[height=0.40in]{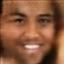} &
		\includegraphics[height=0.40in]{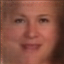} &
		\includegraphics[height=0.40in]{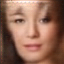} &
		\includegraphics[height=0.40in]{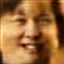} &
		\includegraphics[height=0.40in]{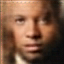} \\~\\~\\

		\textbf{Affine} + \textbf{Integral}: & & & & & & & & \\~\\
		(d)-1 reconstruction &
		\includegraphics[height=0.40in]{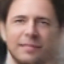} &
		\includegraphics[height=0.40in]{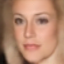} &
		\includegraphics[height=0.40in]{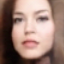} &
		\includegraphics[height=0.40in]{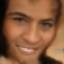} &
		\includegraphics[height=0.40in]{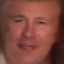} &
		\includegraphics[height=0.40in]{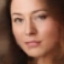} &
		\includegraphics[height=0.40in]{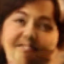} &
		\includegraphics[height=0.40in]{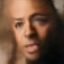} \\    
		(d)-2 texture &
		\includegraphics[height=0.40in]{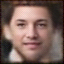} &
		\includegraphics[height=0.40in]{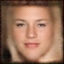} &
		\includegraphics[height=0.40in]{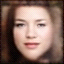} &
		\includegraphics[height=0.40in]{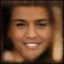} &
		\includegraphics[height=0.40in]{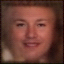} &
		\includegraphics[height=0.40in]{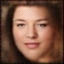} &
		\includegraphics[height=0.40in]{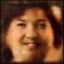} &
		\includegraphics[height=0.40in]{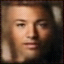} \\
			
		\end{tabular}
	\end{center}
	\caption{Effect of affine and integral warping modules using in our network, using faces in-the-wild. The affine transformation can handle global pose variation, as shown in (b) but not local non-rigid deformation- eyes, noses, or other landmarks are not aligned in the decoded texture images. The proposed integral warping module aligns the faces in a non-rigid manner (c), but in an  exaggerated manner, causing smears in the texture image, e.g. around eyebrows. Incorporating both deformation modules improves the non-rigid alignment (d). In this experiment, we set $Z_A=32$, $Z_T=32$ and $Z_W=32$.}
	\label{fig:supp_affint}
\end{figure}

We visualize the utility of \textit{affine} and \textit{integral} warping modules in our network with face images (Figure \ref{fig:supp_affint}). We can see that the affine transformation handles global pose variance (Figure.~\ref{fig:supp_affint}-(b)) but not local non-rigid deformation. Our proposed integral warping module aligns the faces in a non-rigid manner (Figure \ref{fig:supp_affint}-(c)). Incorporating both deformation modules improves the non-rigid alignment (Figure \ref{fig:supp_affint}-(d)).

\FloatBarrier


\section{Latent Manifold Traversal}

\begin{figure}[ht]
	\centering
	\includegraphics[height=4in]{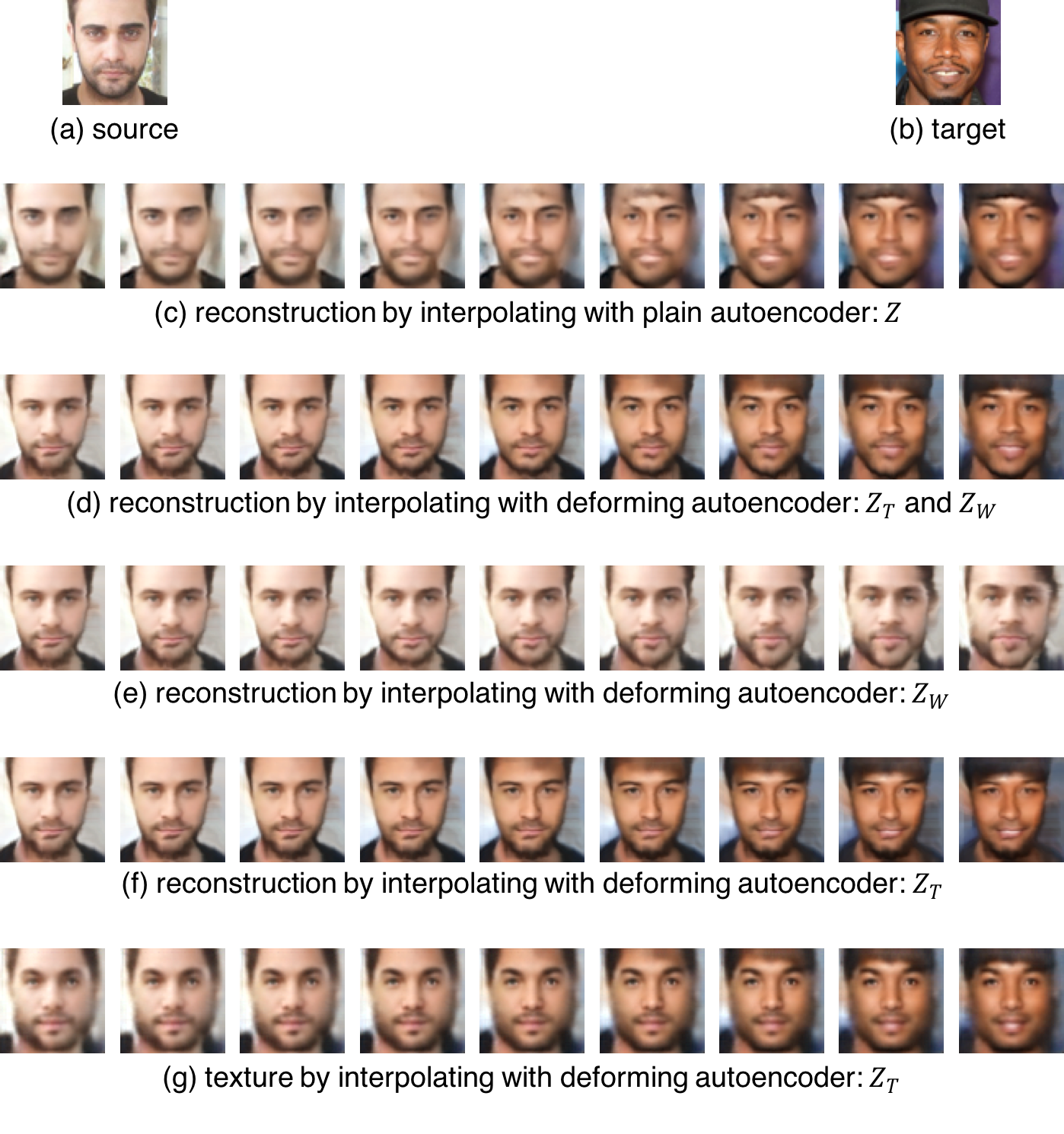}
	\caption{Interpolating learned representations using networks learned on MAFL dataset. Deforming autoencoder learns better latent representations for face compared to a plain autoencoder. By interpolating the latent representations $Z_T$ and/or $Z_W$, we observe smooth transition of pose, shape and skin texture. Interpolated results also stays on the face manifold and, generates more realistic image compared to a plain autoencoder.}
	\label{fig:supp_walk1}
\end{figure}

\begin{figure}[ht]
	\centering
	\includegraphics[height=4in]{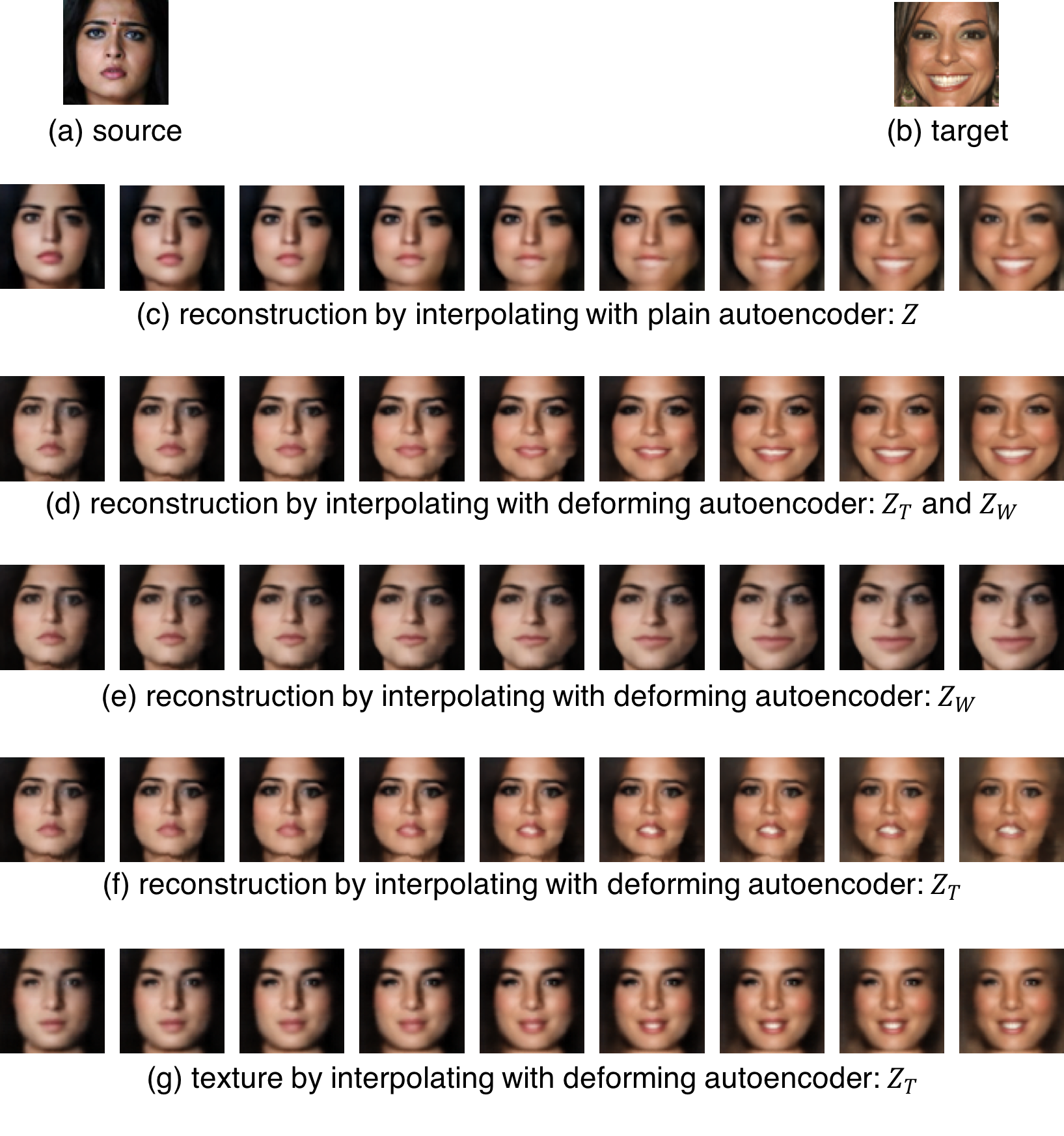}
	\caption{Interpolating learned representations using networks learned on MAFL dataset. Deforming autoencoder learns better latent representations for face compared to a plain autoencoder. By interpolating the latent representations $Z_T$ and/or $Z_W$, we observe smooth transition of pose, shape and skin texture. Interpolated results also stays on the face manifold and, generates more realistic image compared to a plain autoencoder.}
	\label{fig:supp_walk2}
\end{figure}

We provide additional results and comparisons with a plain autoencoder on traversing the learned manifolds. 
In addition to Figure 13 in our manuscript, we provide two more sets of results in Figure \ref{fig:supp_walk1} and Figure \ref{fig:supp_walk2}. Compared to a plain autoencoder, our deforming autoencoder not only generates better reconstructions, but also learns a better  face manifold - interpolating between learned latent representations generates sharper and more realistic face images.
For this experiment, we use the convolutional encoder and decoder architecture as described in Sec. \ref{sec:convnet}.

\begin{figure}[ht]
	\centering
	\includegraphics[height=4in]{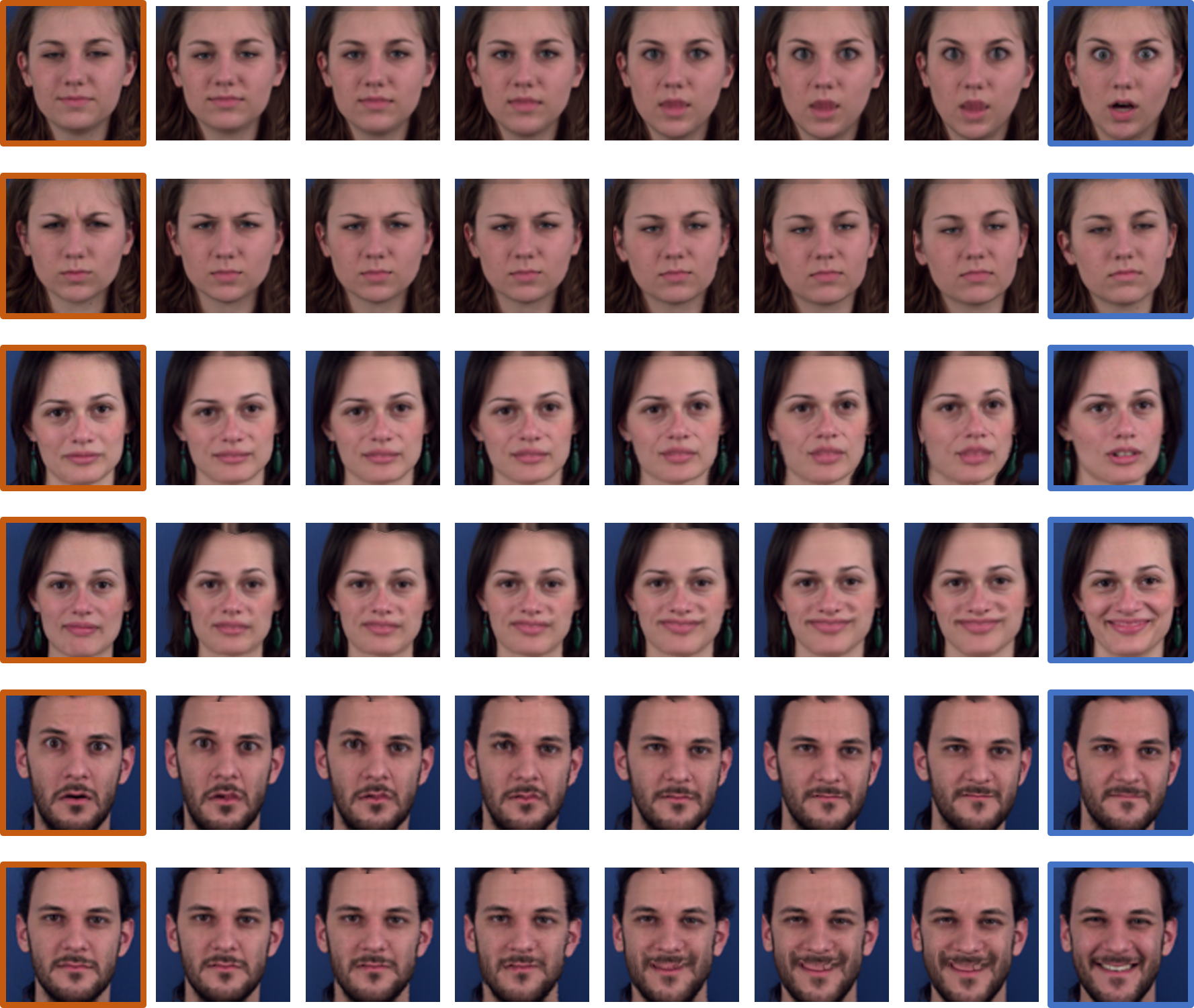}
	\caption{Expression interpolation: Trained on the MUG facial expression dataset, our network is able to disentangle the facial expression
deformation and encode this information in a meaningful latent representation. By interpolating the latent deformation representation
from the source (in orange) to the target (in blue), our network generates sharp images and smooth deformation interpolation between
expressions as shown in each row. In this experiment, ther model for each subject is independently trained, where we set dimension of $Z_T$ to $0$ (assuming single texture for each subject) and dimension of $Z_W$ to $128$.  }
	\label{fig:supp_interp}
\end{figure}
\FloatBarrier

\section{Intrinsic Decomposition with DAE}

\begin{figure}[ht]
	\centering
	\includegraphics[height=6in]{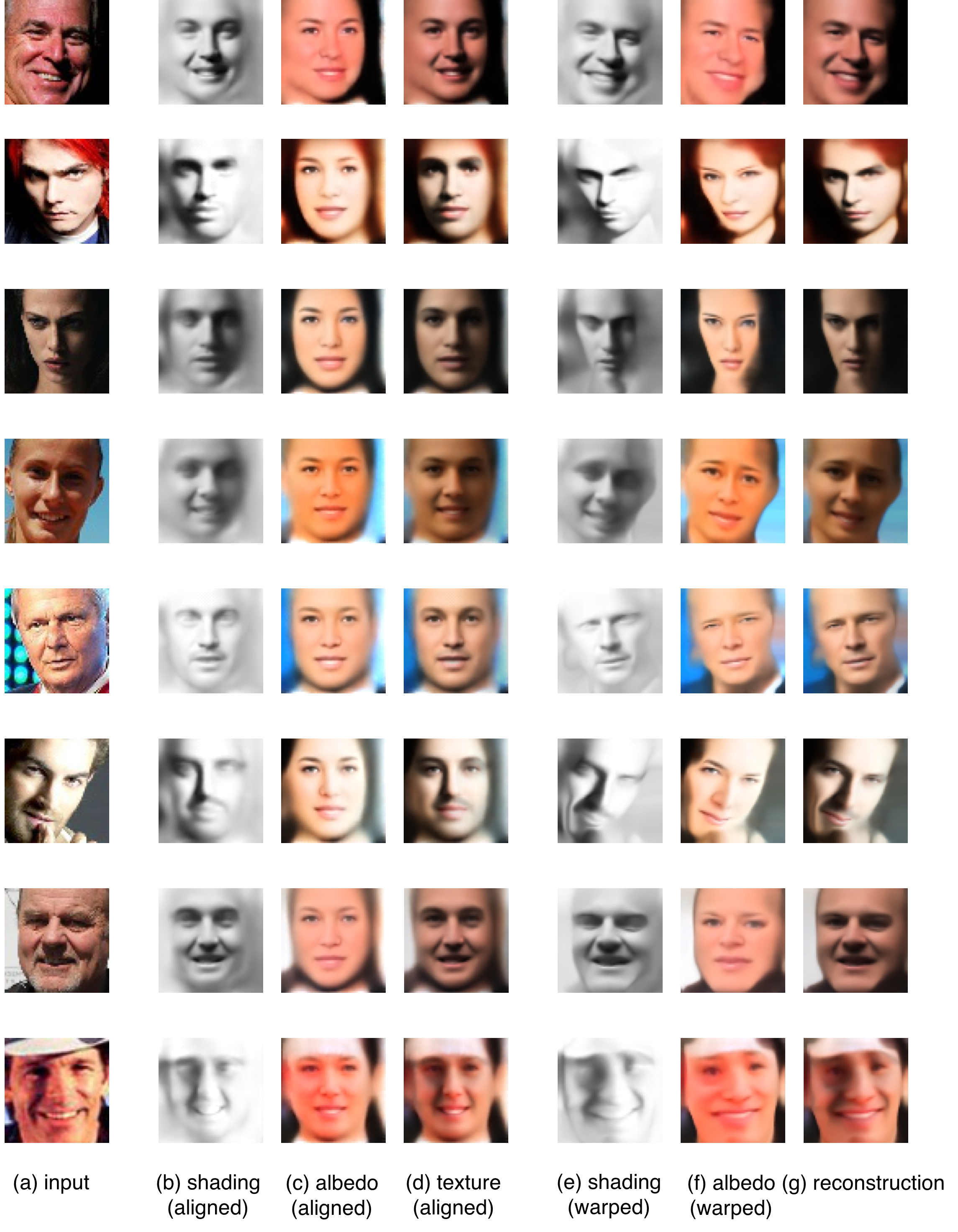}
	\caption{Unsupervised intrinsic decomposition of faces-in-the-wild using Intrinsic-DAE: The network learns to bring faces under different poses and illumination conditions (a) to a canonical view (d), and further separate the shading (b) and albedo (c) component in the canonical view using two independent decoders. With the learned deformation from the deformation decoder we can warp the aligned shading and aligned albedo to its original view as in the input image (e,f).}
	\label{fig:supp_intrinsic}
\end{figure}

In Fig.\ref{fig:supp_intrinsic} we provide additional results  of unsupervised intrinsic disentangling for faces-in-the-wild using Intrinsic-DAE. Using the architecture and objective functions described in Sec. 2.3 of the main paper the network learns to bring faces under different poses and illumination conditions, shown in Fig. \ref{fig:supp_intrinsic}-(a), to a canonical view, as shown in Fig. \ref{fig:supp_intrinsic}-(d), while separating the shading, shown in Fig. \ref{fig:supp_intrinsic}-(b) and albedo, shown in Fig. \ref{fig:supp_intrinsic}-(c) components in the canonical view using two independent decoders. With the learned deformation from the deformation decoder, we can warp the aligned shading and aligned albedo to its original view as in the input image, as shown in Fig. \ref{fig:supp_intrinsic}-(e,f). 

\begin{figure}[ht]
	\centering
	\includegraphics[height=5.2in]{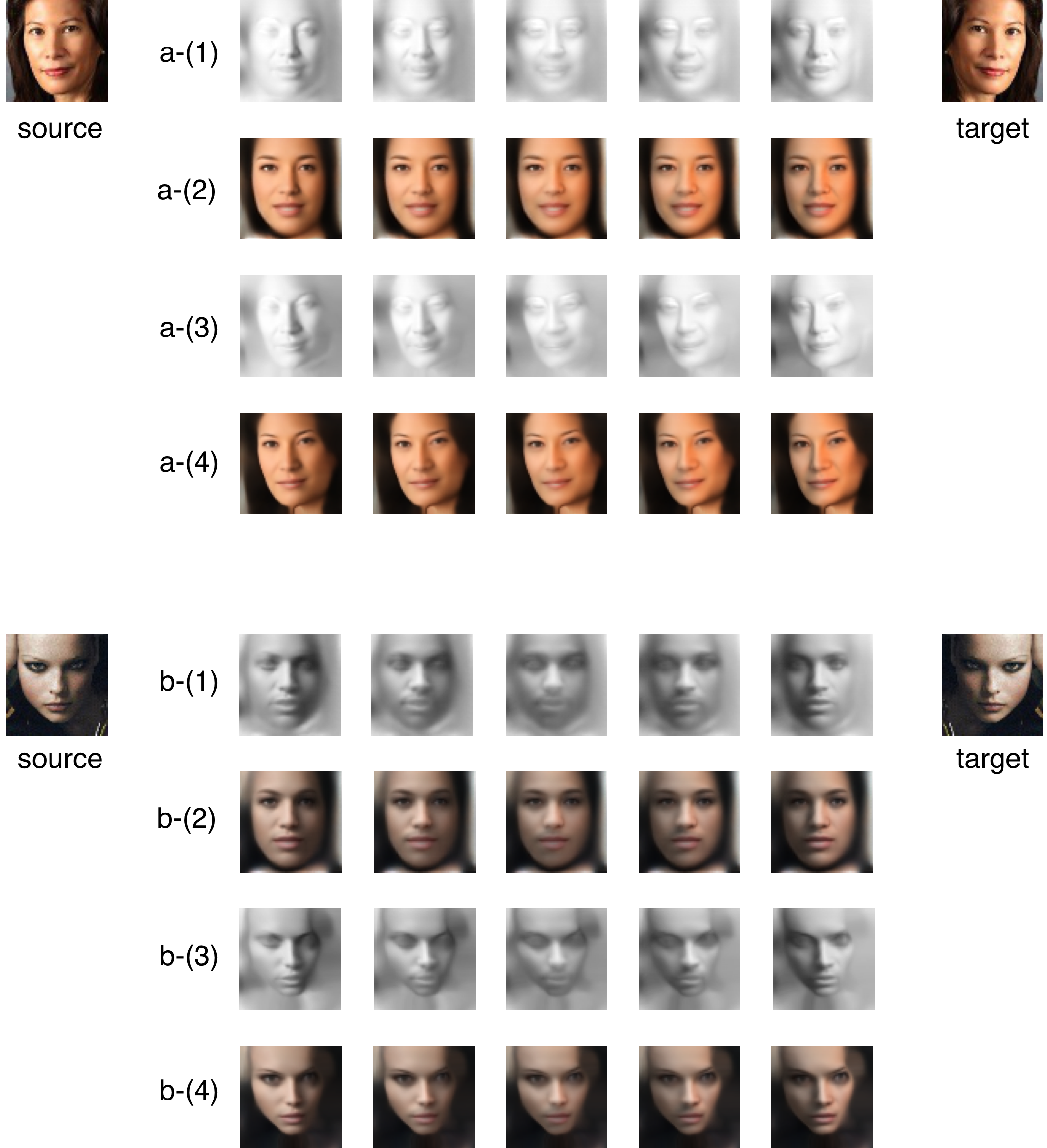}
	\caption{Lighting manipulation by interpolating latent representation of shading: Intrinsic-DAE allows us to disentangle a latent representation for shading for a given face image in an unsupervised manner. Therefore, manipulating the shading component will result in  lighting effects in the output images. In this experiment, we interpolate the latent representation of shading from source to target, which is the mirror of the source with reversed lighting direction. In the result, we can observe that, even without explicitly modeling geometry in our network, we can simulate smooth lighting direction change in both the shading (a-(3), b-(3)) and the final reconstruction (a-(4), b-(4)).  }
	\label{fig:supp_fliplight}
\end{figure}

In Fig. \ref{fig:supp_fliplight}, we provide additional results for ``changing lighting direction'' of a face image using Intinsic-DAE. We show that even without explicitly modeling of geometry, we can simulate smooth and reasonable lighting direction changes in the image by interpolating the learned latent representation for shading, as shown in Fig. \ref{fig:supp_fliplight}-a-(4),b-(4).

For Intrinsic-DAE, we use the DenseNet architecture as the encoders and decoders (\ref{sec:densenet}). The network is trained with a subset of $200,000$ images in the CelebA dataset. The dimensions of latent representations are: 16 for albedo, 16 for shading, and 128 for deformation field.
 
\FloatBarrier

\bibliographystyle{splncs}

\end{document}